\documentclass{article}

\usepackage[final]{neurips_2023}
\usepackage{multirow}
\usepackage{makecell}

\usepackage[utf8]{inputenc} % allow utf-8 input
\usepackage[T1]{fontenc}    % use 8-bit T1 fonts
\usepackage{hyperref}       % hyperlinks
\usepackage{url}            % simple URL typesetting
\usepackage{booktabs}       % professional-quality tables
\usepackage{amsfonts}       % blackboard math symbols
\usepackage{nicefrac}       % compact symbols for 1/2, etc.
\usepackage{microtype}      % microtypography
\usepackage{xcolor}         % colors
\usepackage{subfigure}
\usepackage{graphicx}
\usepackage{algorithm}
\usepackage{algpseudocode}
\usepackage{amsmath}
\usepackage{authblk}
\title{Improving Weak-to-Strong Generalization with \\Scalable Oversight and Ensemble Learning}

\author{\textbf{Jitao Sang}\thanks{These authors contributed equally to this work.}\hspace{1.2mm}\thanks{Corresponding author: jtsang@bjtu.edu.cn}}
\author{\textbf{Yuhang Wang}\protect\footnotemark[1]}
\author{\textbf{Jing Zhang}}
\author{\textbf{Yanxu Zhu}}
\author{\textbf{Chao Kong}}
\author{\\\textbf{Junhong Ye}}
\author{\textbf{Shuyu Wei}}
\author{\textbf{Jinlin Xiao}}
\affil{ Department of Computer Science \authorcr 
Beijing Jiaotong University\authorcr 
Beijing, China 100044}

\begin{document}

\maketitle
\setcounter{footnote}{0}
\renewcommand{\thefootnote}{\arabic{footnote}}

\begin{abstract}
  This paper presents a follow-up study to OpenAI's recent superalignment work on Weak-to-Strong Generalization (W2SG). Superalignment focuses on ensuring that high-level AI systems remain consistent with human values and intentions when dealing with complex, high-risk tasks. The W2SG framework has opened new possibilities for empirical research in this evolving field. 

  Our study simulates two phases of superalignment under the W2SG framework: the development of general superhuman models and the progression towards superintelligence. In the first phase, based on human supervision, the quality of weak supervision is enhanced through a combination of scalable oversight and ensemble learning, reducing the capability gap between weak teachers and strong students. In the second phase, an automatic alignment evaluator is employed as the weak supervisor. By recursively updating this auto aligner, the capabilities of the weak teacher models are synchronously enhanced, achieving weak-to-strong supervision over stronger student models.

We also provide an initial validation of the proposed approach for the first phase. Using the SciQ task as example, we explore ensemble learning for weak teacher models through bagging and boosting. Scalable oversight is explored through two auxiliary settings: human-AI interaction and AI-AI debate. Additionally, the paper discusses the impact of improved weak supervision on enhancing weak-to-strong generalization based on in-context learning. 
Experiment code and dataset will be released at \url{https://github.com/ADaM-BJTU/W2SG}.
\end{abstract}

\section{Introduction}
\begin{figure}[t]
\centering
\subfigure[Evaluated by RLHF.]{
\begin{minipage}[t]{0.86\linewidth}
\centering
\includegraphics[width=\linewidth]{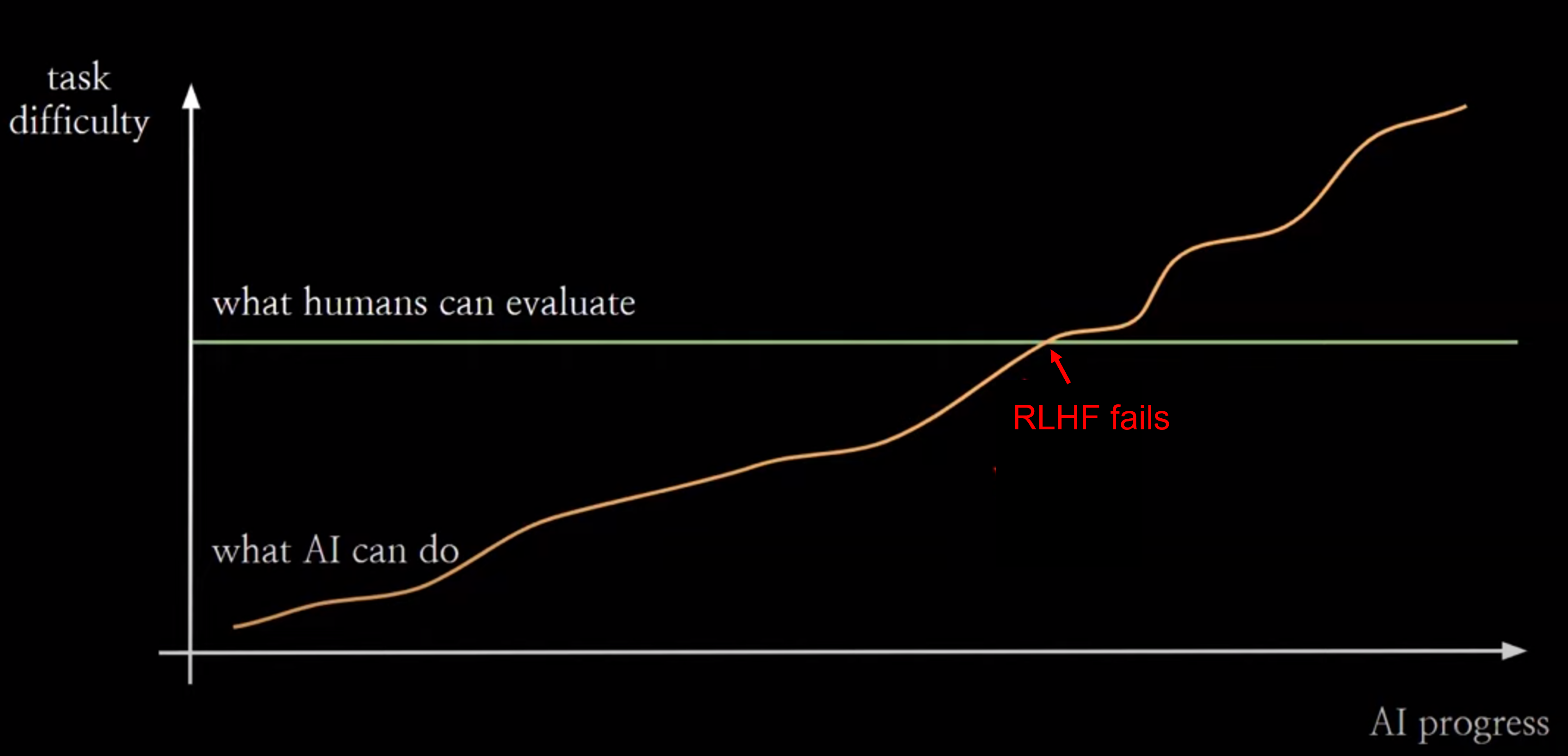}
% \caption{fig1}
\end{minipage}%
}
\quad
\subfigure[Evaluated by SO, W2SG and mW2SG+SO.]{
\begin{minipage}[t]{0.86\linewidth}
\centering
\includegraphics[width=\linewidth]{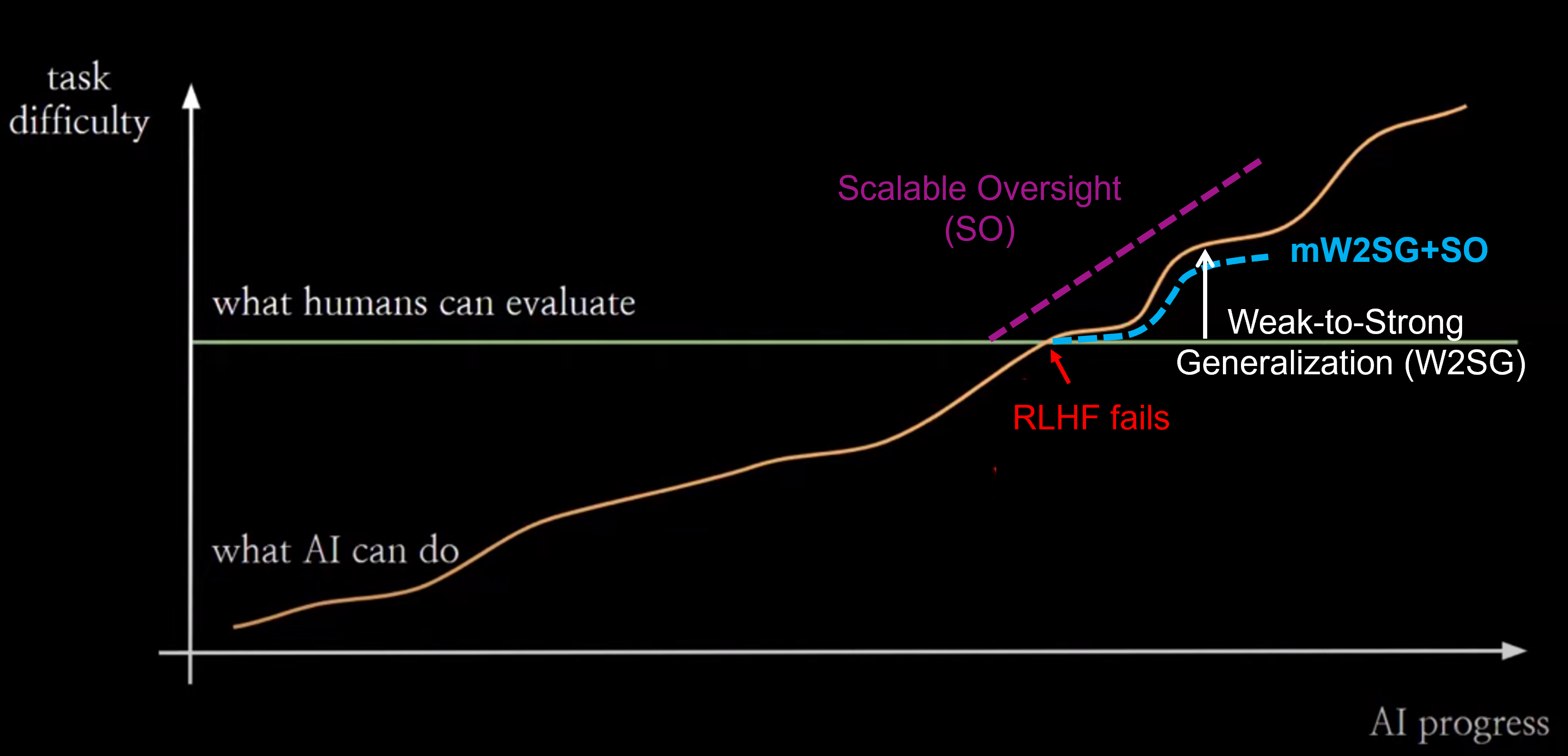}
%\caption{fig2}
\end{minipage}%
}%
\centering
\caption{Task difficulty: AI can do vs. overseer can evaluate .}
\label{fig:1}
\end{figure}

Reinforcement Learning from Human Feedback (RLHF) is the pivotal solution today in aligning powerful AI models with human intentions and values. It has demonstrated considerable success in aligning advanced language models such as GPT-3.5, GPT-4, and Llama2. RLHF is underpinned by the principle of rewarding behaviors highly regarded by humans while penalizing those deemed inferior, as outlined in seminal works~\cite{christiano2017deep, ouyang2022training}. Therefore, its efficacy hinges on the availability of high-quality feedback from human evaluators. As depicted in Figure~\ref{fig:1}(a)~\footnote{Copied from a slide in Jan Leike's talk ``Language and Human Alignment'' (Jan.17, 2023).}, the continual enhancement of AI capabilities presents a conspicuous dichotomy: while AI capabilities are on an upward trajectory, human evaluative capacity remains relatively unchanged. Consequently, we will reach an inflection point where models transcending human-level performance begin to manifest complex behaviors beyond human comprehension. At this juncture, the human ability to provide effective feedback diminishes, and thus RLHF will fail in aligning these superhuman AI models.

One approach to address the above \emph{Superalignment}~\footnote{\url{https://openai.com/blog/introducing-superalignment/}} challenge is \emph{Scalable Oversight} (SO), which seeks to augment the human supervision with carefully designed AI assistants~\cite{bowman2022measuring}. The core idea is to create a composite overseer, that surpasses the capabilities of the AI models being monitored (illustrated by the purple dash line in Figure~\ref{fig:1}(b)). However, this approach encounters a major problem stemming from its overly optimistic assumption: as the AI models grow stronger and more complex, it becomes increasingly challenging to ensure that this augmented human supervision can consistently outpace the AI's advancements.

OpenAI's Superalignment team has recently introduced its first study named \emph{Weak-to-Strong Generalization} (W2SG~\cite{burns2023weaktostrong}). Instead of enhancing the supervisory quality, W2SG leverages weaker models as proxies for human supervisors and aims to ensure that stronger models can generalize from the imperfect supervisory signals of the weaker models. This idea is illustrated by the white arrow in Figure~\ref{fig:1}(b)). W2SG offers a novel perspective to address superalignment challenges and makes empirical research in this area feasible through the analogy between human evaluator and weaker models. It provides a promising avenue by leveraging the nuanced understanding that even less sophisticated AI models can contribute meaningfully to guiding and shaping more advanced models.

From the experimental results in the W2SG paper~\cite{burns2023weaktostrong}, it is easy to find that enhancing the capabilities of the weak teacher models plays a crucial role in improving generalization performance. On the other hand, merely increasing the capabilities of the strong student models, while keeping the weak teacher models unchanged, yields trivial improvements in generalization. This can be intuitively understood: enhancing the weak teacher models' capabilities provides more effective supervisory signals. In contrast, simply boosting the strong student models amplifies the capability gap between the two, thereby escalating the challenges in weak-to-strong generalization.

Therefore, this study is motivated to enhance the performance of weak-to-strong generalization by improving the accuracy of the supervisory signals from the weak teacher models. Specifically, we employ a combination of ensemble learning and scalable oversight approaches: ensemble learning integrates multiple weak teacher models in both serial and parallel schemes to enhance the quality of supervision (multiple Weak-to-Strong Generalization, abbreviated as mW2SG and detailed in Section~\ref{sec-3}), and scalable oversight allows the strong models to assist the weak teacher models in evaluation (abbreviated as SO, detailed in Section~\ref{sec-4}). These two approaches can reduce the capability gap between the weak teacher and the strong student (illustrated by the blue dash line in Figure~\ref{fig:1}(b)), thereby more effectively facilitating weak-to-strong generalization in harnessing the potential of weak supervision to elicit stronger models.

The rest of the paper is organized as follows: In Section~\ref{sec-2}, we introduce a two-phase superalignment roadmap based on the weak-to-strong generalization framework. The first phase is grounded in human supervision, while the second phase pivots to an automated aligned evaluator. Targeting the first phase, Section~\ref{sec-3} and Section~\ref{sec-4} elaborate two respective methods of ensemble learning and scalable oversight to improve weak-to-strong generalization. Experiments and results are also reported. In Section~\ref{sec-5}, we discuss the alternative weak-to-strong generalization setup under in-context learning. The role of scalable oversight in improving weak supervision and contextual example selection is examined.

\begin{figure}[t]
\centering
\subfigure[Human supervision reduces to insignificance.]{
\begin{minipage}[t]{0.86\linewidth}
\centering
\includegraphics[width=\linewidth]{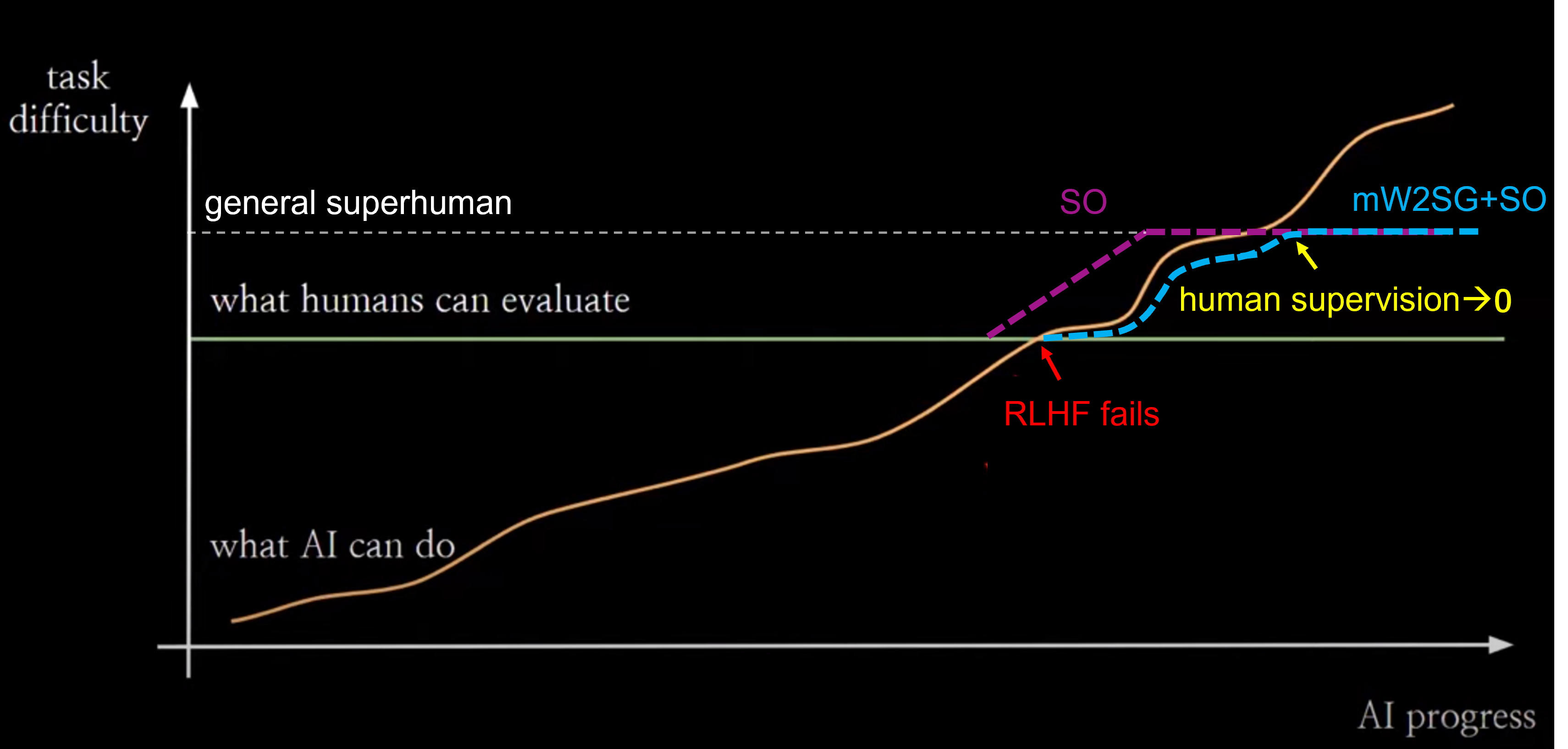}
% \caption{fig1}
\end{minipage}%
}
\quad
\subfigure[Continuously enhancing weak supervision by mW2SG+SO and R-W2SG.]{
\begin{minipage}[t]{0.86\linewidth}
\centering
\includegraphics[width=\linewidth]{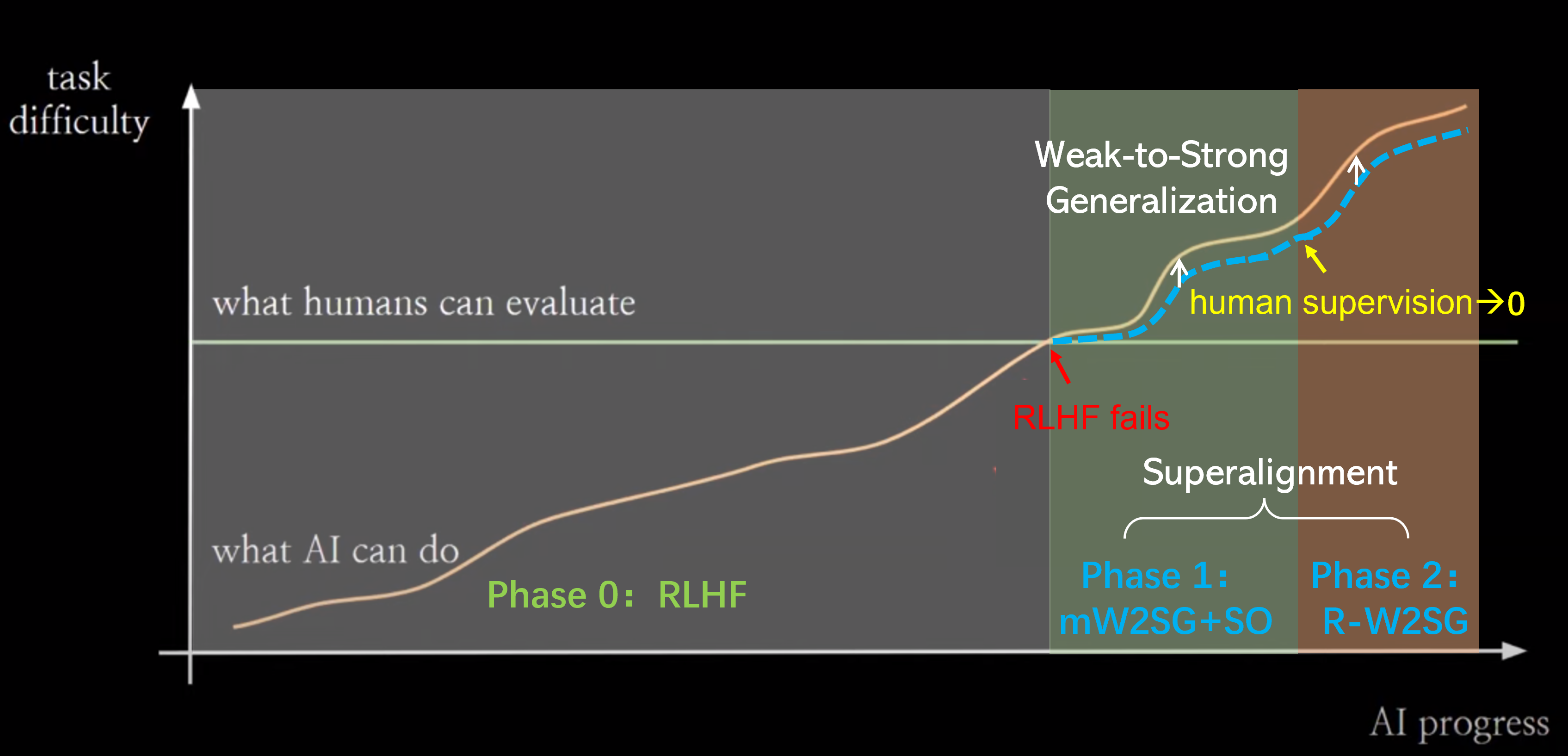}
%\caption{fig2}
\end{minipage}%
}%
\centering
\caption{The roadmap of superalignment via weak-to-strong generalization.}
\label{fig:2}
\end{figure}

\section{The Roadmap to Aligned Super Intelligence via Weak-to-Strong Generalization}
\label{sec-2}
According to the conceptual framework of OpenAI, the realization of superintelligence is envisaged to unfold in two phases: the general superhuman model and superintelligence. Addressing these two pivotal phases, this section will outline the superalignment roadmap under the weak-to-strong generalization setting.

\subsection{Phase 1: Towards General Superhuman Model} 
This initial phase is characterized by the development of AI models that outperform human capabilities in a broad range of tasks. Since the capability gap between the model to be aligned and humans for this phase is not vast, human evaluators can serve as weak supervisors. The approaches proposed in this work, combining ensemble learning and scalable oversight, are specifically targeted at this phase. They aim to enhance the supervisory quality of human evaluators over complex tasks, by integrating multiple weak supervisions and employing assistive AI models.

\subsection{Phase 2: Towards Superintelligence} 
In this advanced phase, the AI models evolve to a level of computational and cognitive capability several orders of magnitude beyond the human level. As the complexity of tasks that AI can handle substantially exceeds human understanding, the supervisory role of human evaluators, even with the assistance of strong auxiliary models, diminishes progressively. As illustrated in Figure~\ref{fig:2}(a), the effectiveness of human supervision reduces to insignificance once student models' capability reaches a certain threshold (e.g., general superhuman), and the evaluative capacity for alignment supervision (represented by the blue dashed line) will remain largely constant.

At this juncture, we propose to replace human evaluators with the strongest aligned student model to serve as the automated alignment evaluator, a new form of weak teacher model, to continue supervising even stronger student models. Notably, unlike the fixed supervisory capacity of human evaluators, the automated alignment evaluator can undergo recursive updates: use the aligned strong student models to update the automated alignment evaluator and facilitate the next generation of weak-to-strong generalization. We call this method \emph{Recursive Weak-to-Strong Generalization} (R-W2SG). As depicted in Figure~\ref{fig:2}(b), the capability of the weak teacher model will increase in sync with the enhancements of the strong student model, ensuring only a capability gap of one generation between the weak teacher and strong student model. This recursive approach ensures a continuous and dynamic alignment process, adapting to the evolving capabilities of AI models. 

Note that in simulating the above recursive process, it is necessary to incrementally increase the complexity of the alignment tasks in sync with the enhancing capabilities of the models. Ideally, the chosen tasks should represent the most complex challenges that the current strongest model can tackle. This strategy ensures that the aligned models are capable of safely and reliably executing the most challenging tasks. 

\begin{figure}[t]
    \centering
    \includegraphics[width=0.99 \textwidth]{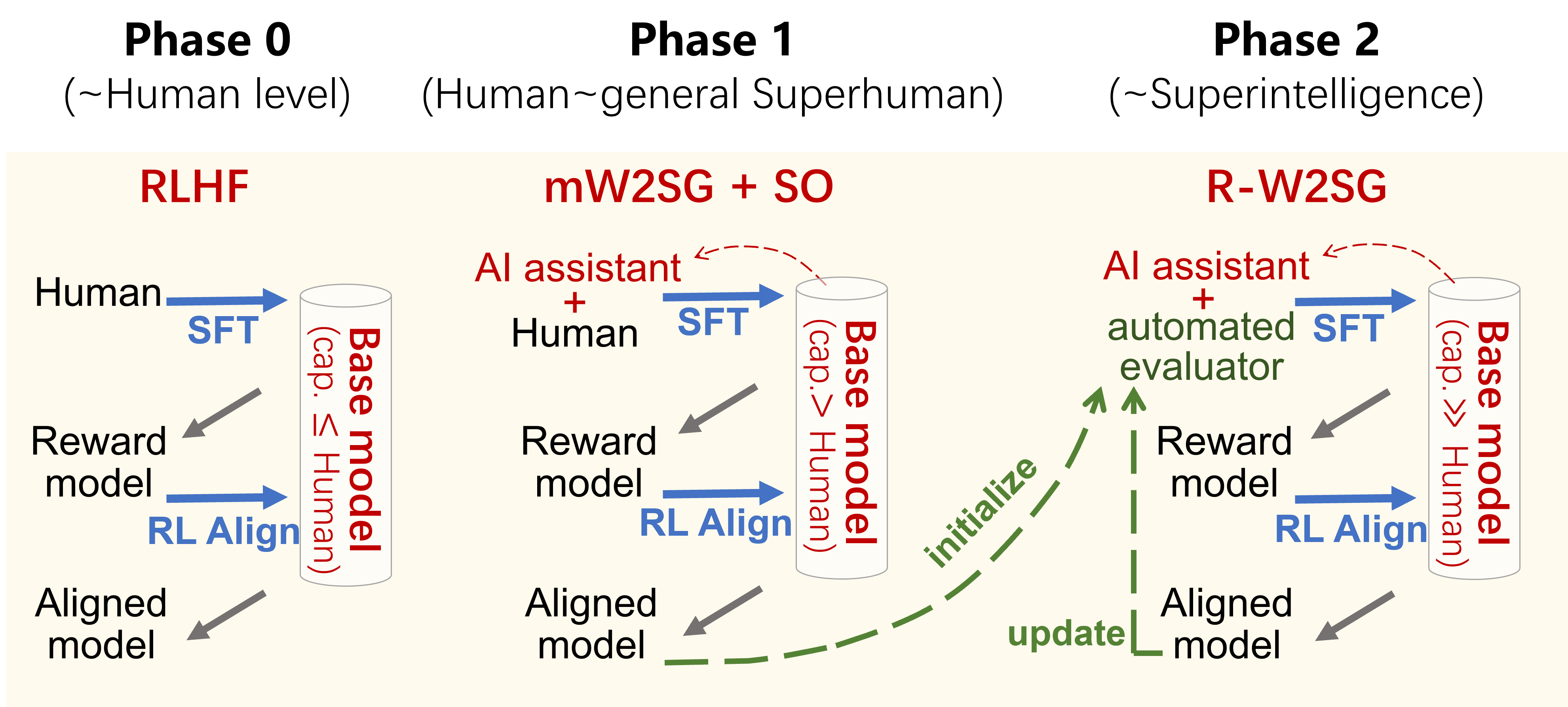}
    \caption{Illustration of the proposed W2SG-based superalignment roadmap.}
    \label{fig:3}
\end{figure}

Figure~\ref{fig:3} summarizes the basic idea of realizing weak-to-strong generalization in the above two phases. This roadmap to aligned superintelligence delineates a strategic and phased solution to developing AI systems that are not only powerful but also remain aligned with human ethics and values. The weak-to-strong generalization framework serves as a foundational basis during this alignment process, ensuring a gradual and controlled evolution toward superintelligence. In the rest of this paper, we focus on Phase 1 to elaborate on the proposed solutions and initial experimental findings.

\section{Improving Weak-to-Strong Generalization with Ensemble Learning}
\label{sec-3}
As an exploratory study, we use the science exam question dataset SciQ~\cite{welbl2017crowdsourcing} as the example task throughout this paper. Originally, the SciQ dataset is structured as a multi-choice format. Strictly following the setup used in OpenAI's paper, we have also converted it into a binary classification problem.  

In terms of model selection, considering the extensive training required for ensemble learning, we opted for the GPT2 series~\cite{radford2019language} in this section. Unless otherwise specified, GPT2-small is fixed as the weak teacher model, and larger-size GPT2-medium, GPT2-large, and GPT2-XL are examined as progressively stronger student models. In the following, we will introduce the methods and experimental results of combining multiple weak teacher models through bagging and boosting, respectively. 

%%军洪，舒羽
\subsection{Bagging-enhanced Weak-to-Strong Generalization}
\vspace{2mm}\textbf{Method}\par \hspace{3mm} 
Bagging is a classic ensemble learning method for enhancing weak model performance by combining multiple weak teacher models in parallel. The key is ensuring the diversity among different weak models. We achieve this by sampling from the training data set and the feature sets. Regarding feature sampling, we can either randomly sample different output feature dimensions, or select different hidden layers as the distinct weak teacher models.

For training set sampling, we employ both random sampling and bootstrap sampling. Random sampling randomly picks data to create a new training set, leaving the rest for validation. Bootstrap sampling, on the other hand, involves repeated selection from the training set, including potential duplicates, with unselected data comprising the validation set. 

For feature sampling, one method is selecting a certain proportion of feature dimensions from the output of the model’s hidden layers. These features are fed into a binary classifier for training. By fixing the sampling proportion and randomly sampling different feature dimensions, different weak teacher models are obtained.

Another method for feature sampling is to choose outputs from different hidden layers of the model. Since each hidden layer processes information differently and captures distinct aspects of the data, it is possible to tap into varying levels of abstraction and complexity in the data representation by choosing outputs from different layers. 

After obtaining multiple different weak teacher models through the above three methods, we have tried both hard voting and soft voting to combine the weak models. In hard voting, the final prediction is the result chosen by the majority of models. In soft voting, models output probabilities for each category as a measure of confidence. The confidence scores are summed and averaged, and the category with the highest average probability is selected as the final prediction. 

\vspace{2mm}\textbf{Experimental Settings} \par
%数据集和实验设置介绍

%数据集：原始的sciq拥有13.7k条数据。我们从原始sciq的训练集采样10k条数据，从这些数据取一半作为弱模型的训练集train1，剩下的数据作为强模型的训练集train2。从剩余的3.7k数据中分别采样1k条作为测试集和验证集
%我们将原始sciq数据集的问题构造为二分类问题，假设每条数据有一个问题和k个候选答案，我们将这条数据转化为k条形式为（Q，A）的k条新数据，其中错误答案A的标签为0，正确答案A的标签为1。在这个过程中我们将保持每个问题正确和错误的标签数量相同，因此每个问题将会有两个（Q,A）。最终train1和train2各拥有10k条数据，test和validation各拥有2k条数据。

%模型：采样训练样本的方式采用gpt2-small和gpt2-medium作为弱老师模型，强模型采用gpt2-medium，gpt2-large和gpt2-xl；由于gpt2-small隐藏层少，并且输出层特征的维度较少，因此采样特征与采样不同隐藏层的方法采用gpt2-medium作为弱模型，强模型采用gpt2-large和gpt2-xl。

%弱标签：我们在train1上训练弱模型，并在train2和validation上进行回答，将模型的输出输入二分类器得到“错误”和“正确”的概率作为弱标签。

%超参数：弱模型的训练的batch_size为32，采样训练样本的方式训练5个epoch，其余两种方式分别训练4个epoch。我们在验证集上监控模型的性能，当模型性能在验证集上不再提升时停止训练。

%训练方式:采样训练样本的方式分别在train1上采样4000、5000、6000、7000、8000和10000（boostrap）条数据训练20个弱模型，其余的数据作为验证集。其中正样本和负样本各占50%。在整体上观察采样训练样本的结果，采样不同隐藏层的方式分别取gpt2-medium的第10-24层隐藏层的输出在train1上训练20个弱模型。采样输出层特征的方式分别随机采样gpt2-medium输出层特征的10%、30%、50%和70%并用这些特征在train1上训练20个弱模型。

% Originally, SciQ has $13.7k$ exam questions.We sample $10k$ data points from the original SciQ, half of which are used as the training set (``train1'') for weak models, while the remaining data serves as the training set (``train2'') for strong models. Additionally, sample 1k respectively from the remaining $3.7k$ data for the test set and the validation set.

% We conduct experiments using the SciQ\textbf{[cite]} dataset. 
\hspace{3mm}\textbf{Dataset.} Initially, we sample $10k$ data points from the training set of SciQ and divide them evenly into two parts. 
Additionally, we extract $1k$ samples from both the validation and test sets of the original SciQ dataset.  These samples are designated as our validation and test sets, respectively.
Furthermore, we convert this dataset into binary classification problems. 
Assuming each data point consists of one question and $k$ candidate answers, we transform each data point into $k$ new data points in the form of $(Q, A)$, where the label for incorrect answer $A$ is 0 and the label for correct answer $A$ is 1. Throughout this process, we ensure an equal number of correct and incorrect labels for each question, resulting in two $(Q, A)$ pairs per question.
Finally, our constructed ``train'' and ``train2'' each contain $10k$ samples, while the test set and validation set each contain $2k$ samples.\par

\hspace{3mm}\textbf{Models.} 
For training set sampling, GPT2-small and GPT-medium are employed as the weak teacher model, and larger-size models are employed as the strong student models (i.e., weak model of GPT-2 small with strong model of GPT2-medium, GPT2-large and GPT2-XL). For feature sampling, due to the fewer hidden layers of the GPT2-small and the smaller dimensions of the output layer features, we only use GPT2-medium as the weak model and GPT2-large and GPT2-XL as the strong model.\par

\hspace{3mm}\textbf{Weak labels.} 
We train weak models on ``train1'' and use them to predict samples on ``train2'' and validation. The outputs of the models are fed into a binary classifier to obtain the probabilities of ``incorrect'' and ``correct'', serving as weak labels.\par

\hspace{3mm}\textbf{Hyperparameters.} 
We set the batch size for training weak models to 32. For the method involving sampled training samples, the models are trained for 5 epochs. For the other two methods, they are trained for 4 epochs each. Model performance is monitored on the validation set, and training is halted when there is no further improvement in performance on the validation set.\par

\hspace{3mm}\textbf{Sampling details.}
Training set sampling strategy involves selecting 4000, 5000, 6000, 7000, 8000, and 10000 (bootstrap) data. The unselected training samples constitute the validation set. Among them, positive and negative samples each account for $50\%$. Feature layer sampling involved randomly sampling 10\%, 30\%, 50\%, and 70\% of features from the GPT2-medium output layer and then using these sampled features to train weak models on ``train1''. Feature dimension sampling involved using outputs from the 10th to 24th hidden layers of the GPT2-medium model to train weak models on ``train1''. 

\begin{figure}[t]
\centering
\subfigure[Soft voting. ]{
\begin{minipage}[t]{0.485\linewidth}
\centering
\includegraphics[width=\linewidth]{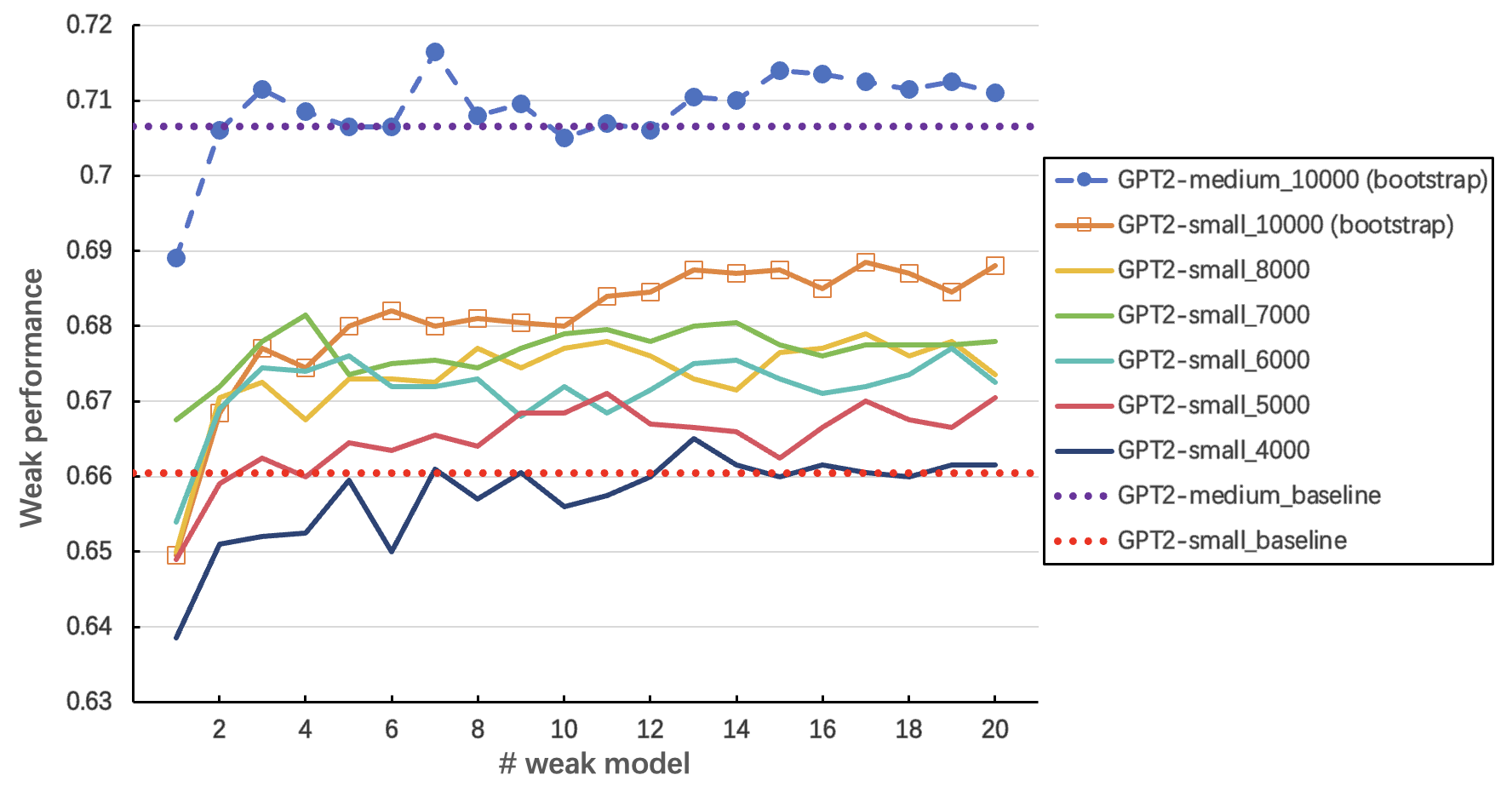}
\label{subfig:4-1}
% \caption{fig1}
\end{minipage}%
}
\subfigure[Hard voting.]{
\begin{minipage}[t]{0.485\linewidth}
\centering
\includegraphics[width=\linewidth]{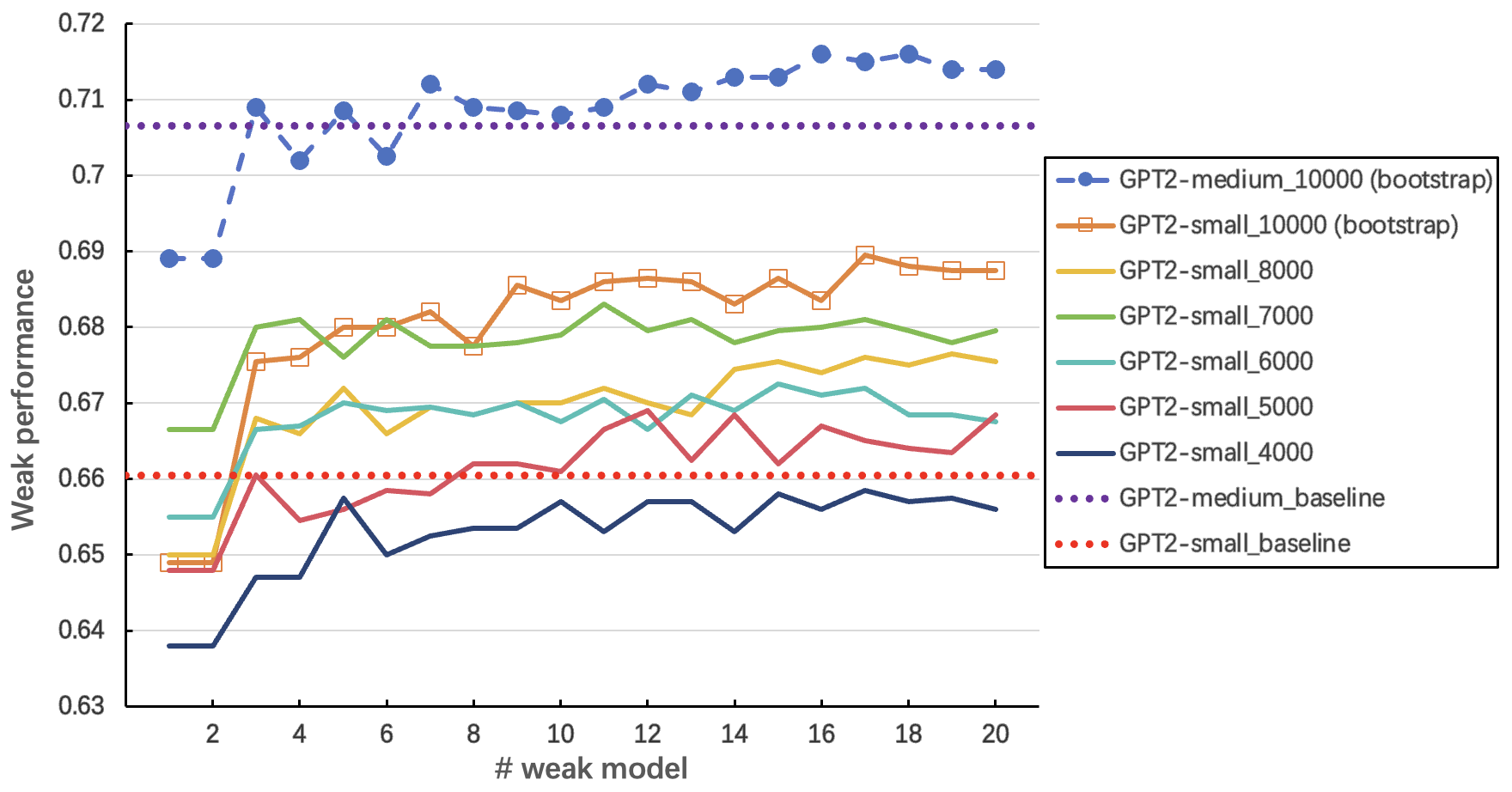}
\label{subfig:4-2}
%\caption{fig2}
\end{minipage}%
}%
% \caption{Weak models with different training sets.}
\caption{Weak models with different training set sampling.}
% \\
% \subfigure[Weak models with different feature dimensions (soft voting).]{
\subfigure[Soft voting.]{
\begin{minipage}[t]{0.485\linewidth}
\centering
\includegraphics[width=\linewidth]{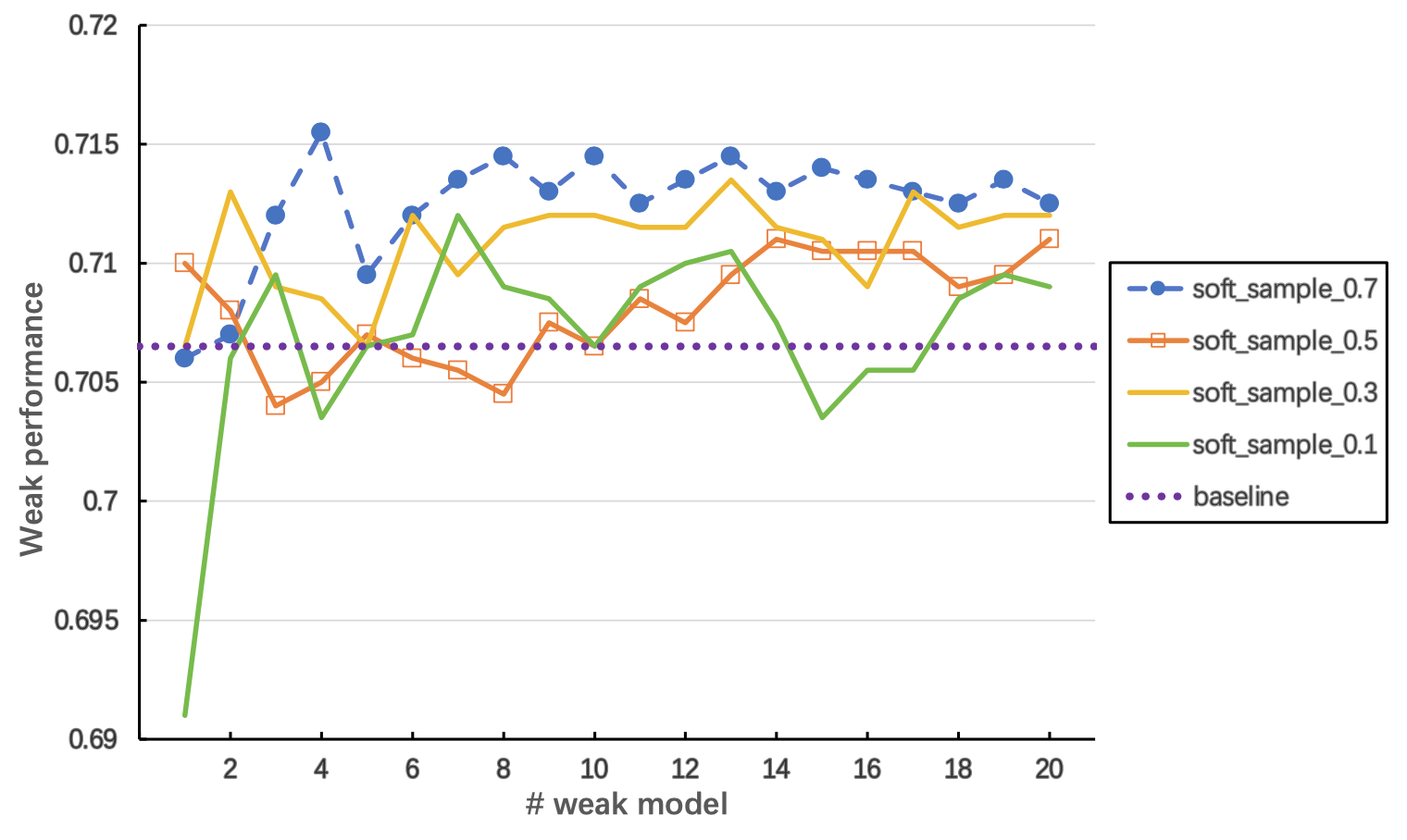}
\label{subfig:4-3}
%\caption{fig3}
\end{minipage}%
}%
% \subfigure[Weak models with different feature dimensions (hard voting).]{
\subfigure[Hard voting.]{
\begin{minipage}[t]{0.485\linewidth}
\centering
\includegraphics[width=\linewidth]{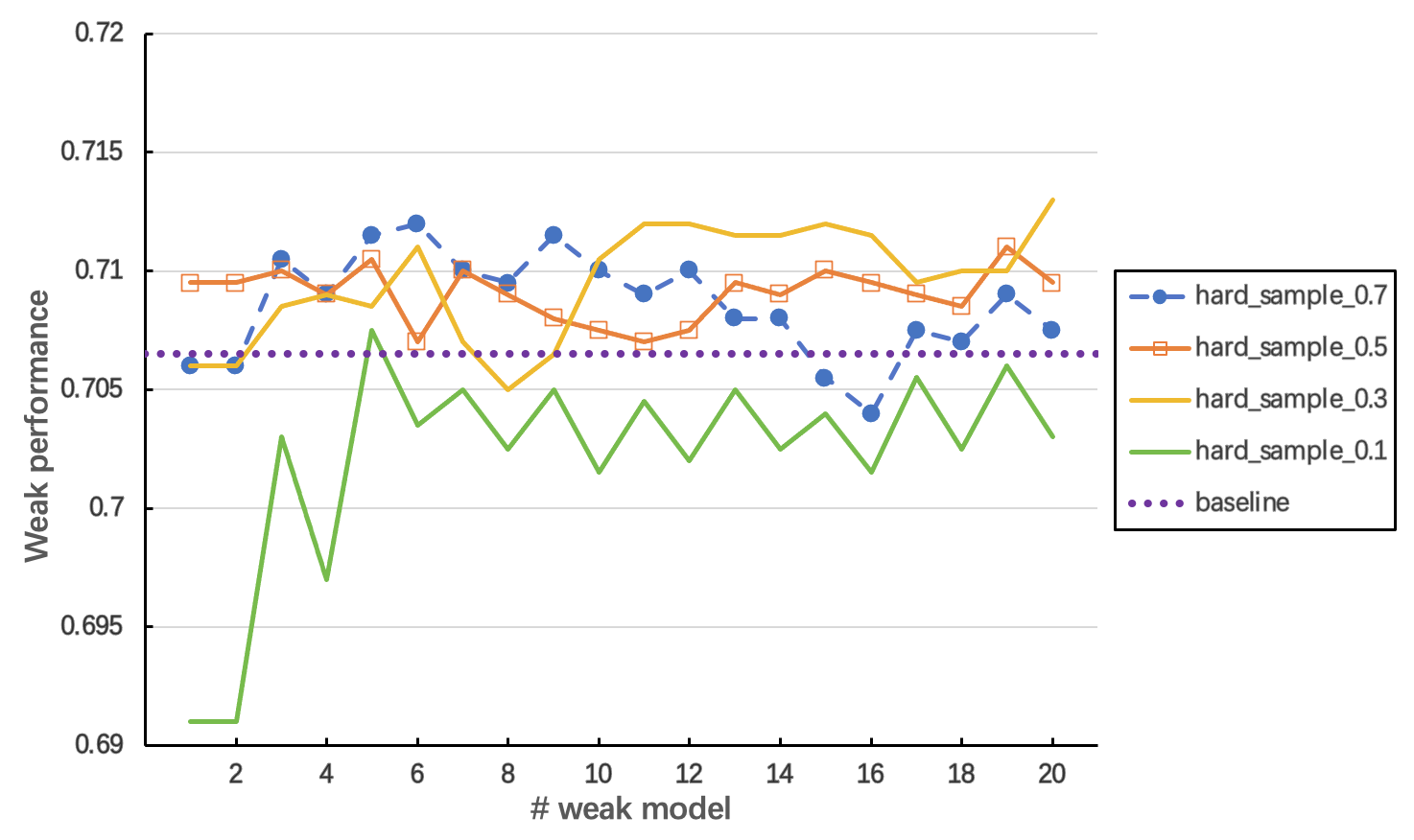}
\label{subfig:4-4}
%\caption{fig4}
\end{minipage}%
}%
\caption{Weak models with different feature set sampling.}
% \\
% \subfigure[Weak models with different feature layers.]{

% \subfigure{
% \begin{minipage}[t]{0.485\linewidth}
% \centering
% \includegraphics[width=\linewidth]{figs/4.5.png}
% \label{subfig:4-5}
% %\caption{fig5}
% \end{minipage}%
% }%
% \caption{Weak models with different feature layer combination.}
% \centering
% \label{fig:4}

\end{figure}

\begin{figure}[t]
    \centering
    \includegraphics[width=0.55 \textwidth]{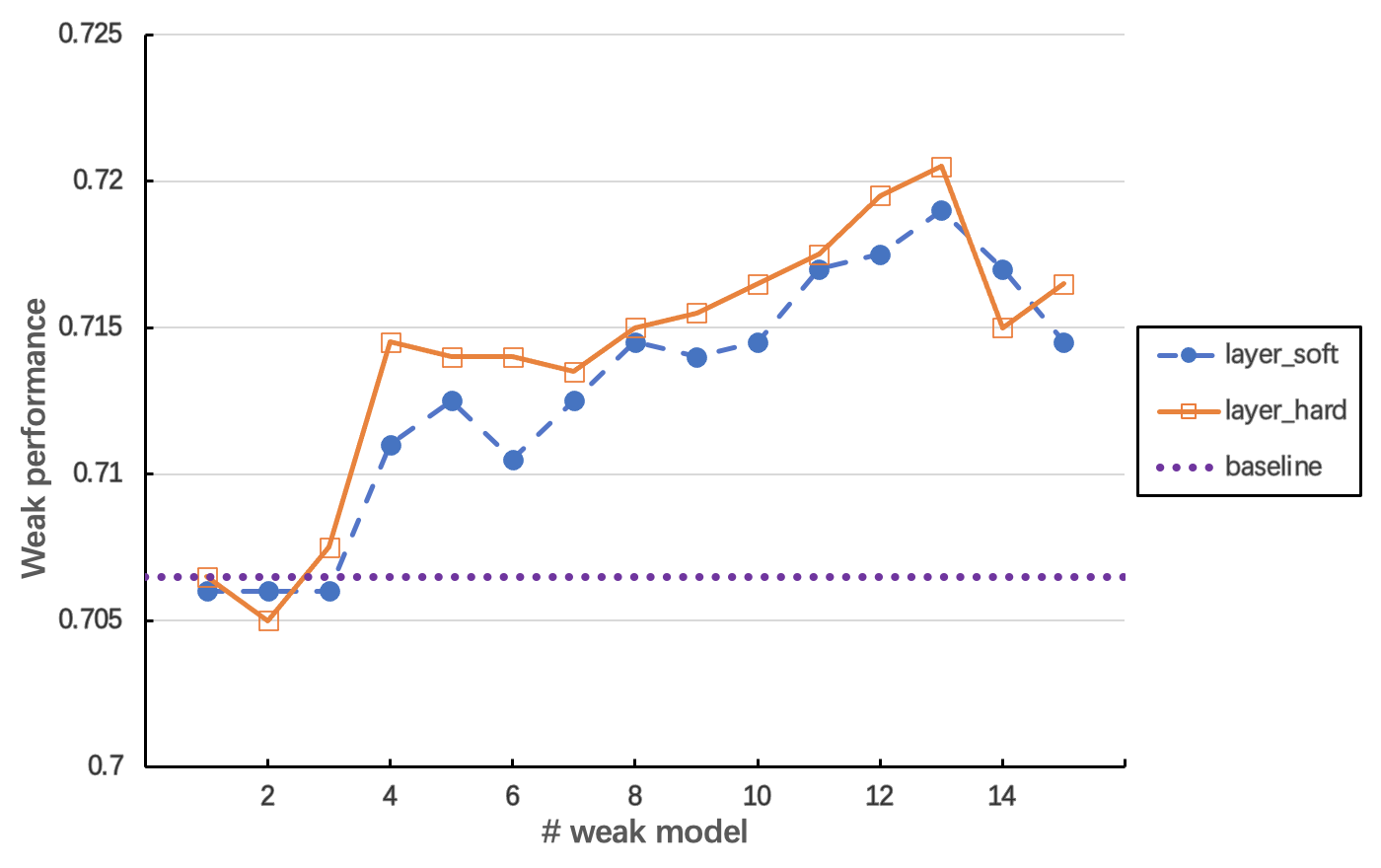}
    \caption{Weak models with different feature layer combination.}
    \label{fig:4.5}
\end{figure}

\vspace{2mm}\textbf{Experimental Results}\par
%当集成的弱模型数量较少时，测试准确率呈线性增长，但随着模型数量的进一步增加，准确率提升逐渐趋于平稳。在训练数据较为有限时，采用随机采样策略的模型集成表现与训练数据数量呈正相关。这是因为随着训练数据增加，弱模型得到更充分的训练。然而，当训练数据达到一定数量（例如3500条）后，这一趋势变得不明显，甚至在硬投票中性能略微下降。这可能是因为随着数据增加，数据的多样性几乎没有变化，导致弱模型的多样性增长缓慢。由于弱模型的多样性是影响集成模型性能的关键因素之一，因此在性能上没有显著增长。与随机采样相比，采用bootstrap策略集成后性能更优，这表明通过这种方式采样的数据可能具有更高的多样性。
%更新：采样数量统一算总数，改成4000、5000、6000、7000、8000、10000（bootstrap）。
We present the results of training set sampling in Figure~\ref{subfig:4-1} and Figure~\ref{subfig:4-2}. As the number of ensembled models increases, there is a clear and linear enhancement in accuracy. The bagging performance when exceeding 5 ensembled models generally surpasses the baseline. However, beyond 10 ensembled models, the increase in accuracy begins to plateau. Regarding the different sizes of the training set, as randomly sampled training set increases from 4000 to 7000 samples, the ensemble performance consistently improves. However, when the training data further increases beyond 7000, this upward trend diminishes, and performance may even slightly decline. This can be attributed to a decrease in data diversity, which in turn diminishes the effectiveness of the ensemble. Comparatively, bootstrap-based ensembles demonstrate the best performance, by sampling as much data as possible while ensuring diversity.

% when the number of weak models in the ensemble is small, the test accuracy shows a linear increase. However, as the number of models further increases, the accuracy improvement gradually tends to stabilize. When the training data is limited, the performance of the model ensemble using random sampling strategy is positively correlated with the number of training data. This is because with the increase of training data, weak models receive more sufficient training. However, when the training data reaches a certain quantity (e.g., 7000 samples), this trend becomes less obvious, and even slightly decreases in performance in hard voting. This may be due to the fact that with the increase of data, the diversity of the data almost does not change, resulting in slow growth of weak model diversity. Since the diversity of weak models is one of the key factors affecting the performance of the ensemble model, there is no significant growth in performance. Compared with random sampling, the performance of ensemble using bootstrap strategy is better, indicating that the data sampled in this way may have higher diversity.

%在整体上观察采样输出层特征的实验结果，采样特征比例为0.1时，弱模型性能受到较大影响，集成5个弱模型后准确率达到最高，随着弱模型个数进一步增加，准确率趋于平稳；随着采样特征比例增加，不同采样特征比例的弱模型性能相差不大，并且集成后模型性能提升较小，可能是因为采样得到的特征同质化严重。 

Figure~\ref{subfig:4-3} and Figure~\ref{subfig:4-4} display the results of sampling different output-layer feature dimensions. Notably, when the feature sampling proportion is set to 0.1, there's a significant impact on the performance of ensembling. With higher feature sampling proportions, the variance in performance among weak models is marginal.

%在整体上观察采样不同隐藏层的实验结果，随着模型个数得增长，测试的准确率呈线性增长趋势，并在集成13个模型时准确率达到最大。这可能是因为弱模型不同层之间存在明显差异，导致训练后的模型具有较高的多样性，因此在一定数量内增加集成的模型能持续提高模型性能。
 The results of combining different hidden layers are shown in Figure~\ref{fig:4.5}. We observe a linear increase with the rising number of ensembled models, peaking at the ensemble of 13 models (i.e., ensembling 10th to 22nd feature layers). This trend is likely due to the substantial differences among the hidden layers of the weak models, which leads to a high degree of diversity in the trained models. Consequently, augmenting the number of models in the ensemble within a certain limit can further enhance the performance.

\begin{table}[t]
\centering
\begin{tabular}{|l|l|>{\centering\arraybackslash}p{1.4cm}|>{\centering\arraybackslash}p{1.4cm}|>{\centering\arraybackslash}p{1.4cm}|} 

\hline  
& \multirow{2}{*}{Weak performance }  & \multicolumn{3}{c|}{Weak-to-Strong performance} \\ 
\cline{3-5} 
 &  &  medium &  large & XL  \\      
\hline 
\multirow{2}{*}{before bagging}  &0.661 (small)& 0.659 & 0.676 & 0.682 \\
                &0.694 (medium)& -- & 0.717 & 0.730 \\
                                         
\hline
\multirow{2}{*}{after bagging (bootstrap sample)}  &0.689 (small) & 0.678 & 0.688 & 0.682\\ 
&0.714 (medium) & -- & \textbf{0.727} & \textbf{0.738}\\
                                 
\hline
\multirow{1}{*}{after bagging (layer combine)}  &\textbf{0.719} (medium)& -- & 0.721 & 0.731\\
\hline
\end{tabular}
\vspace{2mm}
\caption{Improving W2SG via bagging-based supervision enhancement. }
\label{tab: bagging train set}
\end{table}
We also report weak-to-strong performance based on the above ensembled models. It can be observed that bootstrap and feature layer sampling have respectively achieved better improvements in training data sampling and feature sampling. Consequently, we integrate 17 GPT2-small and 7 GPT2-medium models, each trained on 10000 bootstrapped samples, for data-based bagging. For feature-based bagging, we integrate outputs from the 10th to the 22nd hidden layers of GPT2-medium models.

The experimental results are shown in Table \ref{tab: bagging train set}. We find two primary observations: (1) Horizontally, as the size of the strong model escalates, the influence of enhanced weak performance on weak-to-strong generalization gradually diminishes. This may stem from the excessively wide gap between weak and strong models, hindering the weak model's effective supervision over the strong model. This underpins our emphasis in Phase 2 \emph{R-W2SG} on maintaining a manageable capability gap between the weak and strong models within one generation. (2) Vertically, although integrating different hidden layers yields a higher boost in weak performance compared to bootstrap sampling (0.719 vs. 0.714), it does not contribute as significantly to weak-to-strong performance (0.721 vs. 0.727 for GPT2-large and 0.731 vs. 0.738 for GPT2-XL). This may be attributed to bootstrap sampling preserving a more robust diversity among weak models. These two findings indicate that in future studies on weak-to-strong generalization, attention should extend beyond merely enhancing weak supervision, to encompass the influences of other pivotal factors.

\begin{figure}[t]
\centering
\subfigure[Ensemble with AdaBoost.]{
\begin{minipage}[t]{0.485\linewidth}
\centering
\includegraphics[width=\linewidth]{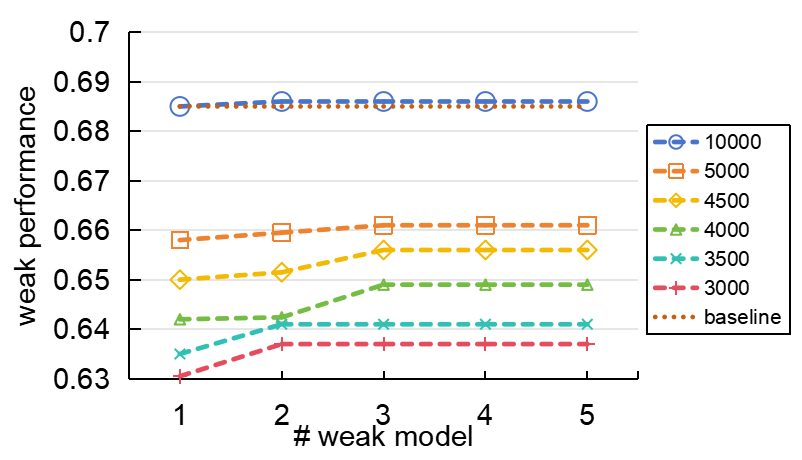}
% \caption{fig1}
\end{minipage}%
}
\subfigure[Ensemble with Gradient Boost.]{
\begin{minipage}[t]{0.485\linewidth}
\centering
\includegraphics[width=\linewidth]{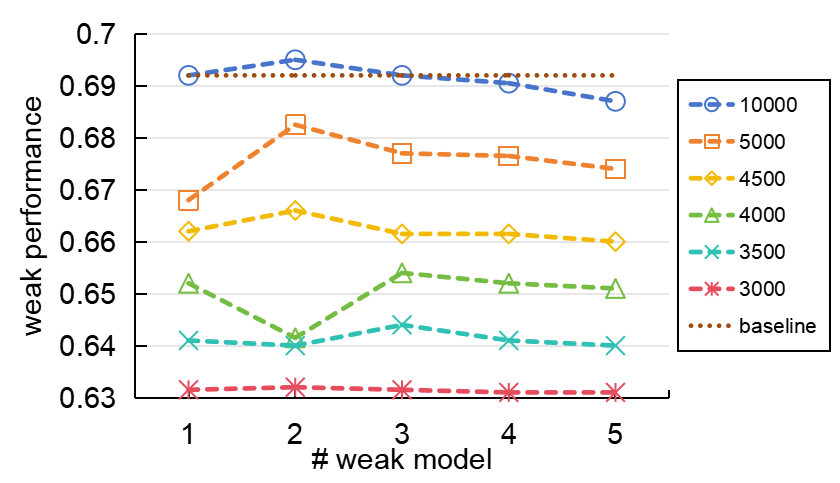}
%\caption{fig2}
\end{minipage}%
}%
\centering
\caption{Weak models with different boosting methods.}
\label{fig:6}
\end{figure}

%%孔超
\subsection{Boosting-enhanced Weak-to-Strong Generalization}
\vspace{2mm}\textbf{Method}\par
%介绍基本思路
Boosting serially trains multiple weak models, where each subsequent model is designed to address and rectify the deficiencies of its predecessors. We implement both AdaBoost and Gradient Boost to combine with Weak-to-Strong generalization.

The core of the Adaboost method is that when the current weak classifier is difficult to learn from some samples, the weight of these difficult samples is increased when the next weak classifier is trained. So the latter weak model can make up for the shortcomings of the previous model and be more comprehensive after integration.

Gradient Boost, on the other hand, guides the update of the current weak classifier in the direction opposite to the gradient of the cumulative previous models. This method effectively performs gradient descent relative to the loss function of the target, making the ensemble model exhibit reduced loss on the training set. The processes of calculating sample weight is listed in Alg.~\ref{alg:adaboost} and Alg.~\ref{alg:gradientboost} in Appendix.\par
%实验设置介绍

\begin{table}[t] 
\centering
\begin{tabular}{|l|c|>{\centering\arraybackslash}p{1.6cm}|>{\centering\arraybackslash}p{1.6cm}|>{\centering\arraybackslash}p{1.6cm}|} 

\hline  
& \multirow{2}{*}{Weak performance }  & \multicolumn{3}{c|}{Weak-to-Strong performance} \\ 
\cline{3-5} 
 &  &  medium &  large & XL  \\      
\hline 
\multirow{1}{*}{before boosting} &0.685 (small)& \textbf{0.672} & 0.680 & \textbf{0.685} \\        
\hline
\multirow{1}{*}{after boosting (Adaboost)} & 0.686 (small) & 0.671 & 0.675 & 0.676\\ 
\hline
\multirow{1}{*}{after boosting (Gradient Boost)}  &\textbf{0.695} (small)& \textbf{0.672} & \textbf{0.688} & 0.679\\
\hline
\end{tabular}
\caption{Improving W2SG via boosting-based supervision enhancement.}
\label{tab: boost train set}
\end{table}

\begin{table}[t]
\centering
\begin{tabular}{|c|c|c|c|c|c|} 
\hline  

Ensemble method &  1st model & 2nd model  & 3rd model  & 4th model & 5th model\\      
\hline 
Adaboost & \textbf{0.685} & 0.638 & 0.52 & - & -\\        
\hline
Gradient Boost & \textbf{0.692} & 0.689 & 0.668 & 0.620 & 0.542 \\ 

\hline
\end{tabular}
\caption{Single weak models' performance in boosting which are trained with 10000 samples. From the third round onwards, the performance of single Adaboost weak model is close to 0.5.}
\label{tab: boosting single weak model}
\end{table}

\vspace{2mm}\textbf{Experimental Results}\par
%实验结果-AdaBoost、Gradient设置下weak model结果
%如图{}的(a)、(b)所示，两个方法集成时都会遭遇明显的天花板效应。在2到3个模型集成时会使性能小幅上升，但更多模型会导致集成失败。
%以adaboost结果为例， 从表{}中可以看出: 
%(1) 在训练集准确率较大的情况下，验证集准确率较低，表明此任务存在泛化困难。本任务的设置为知识问答，存在较多的OOD情况。
%(2) 第一个模型性能显著强于后续训练的模型，甚至串行训练的第三个模型在对训练集也无法有效拟合。所以后续训练的模型即便能预测对某些困难样本，但由于其整体性能的快速下降，对集成性能的提升依然有限。
%我们分析了这些模型性能下降的原因。从图{}可以看出，后续训练的串行模型损失难以下降，表明在训练阶段时模型就产生了崩溃，即在给各个样本重新分配权重后，模型无法捕获数据集里的正确特征。从表{}中可以看出，模型训练崩溃时，高权重样本(>1.2)的权重占比显著上升，即重点在这些困难样本训练后，无法学习到通用特征。考虑到任务为复杂的知识问答，重点关注于其中的难题时，会极大偏离整体特征。
%考虑weak2strong任务会不断迭代，变得越来越复杂，这种单纯提高困难样本权重的训练方法反而不利于整体任务的学习。 

We replicate the experimental setup from the previous subsection for bagging. Ensemble accuracy results for weak models using AdaBoost and Gradient Boost are depicted in Figure~\ref{fig:6}. Notably, comparable outcomes are only achieved when training with the full dataset. While Gradient Boost shows a slight improvement with 2-3 model integrations, further combinations lead to a decline. Table~\ref{tab: boosting single weak model} details the performance of the initial few weak models. It's observed that the first model performs best, with a rapid decrease from the third model onwards, nearing random guess levels. This may be attributed to the limitations of standard ensemble methods in generative problems and integrating large-sized models.

\begin{figure}[t]
    \centering
    \includegraphics[width=0.6 \textwidth]{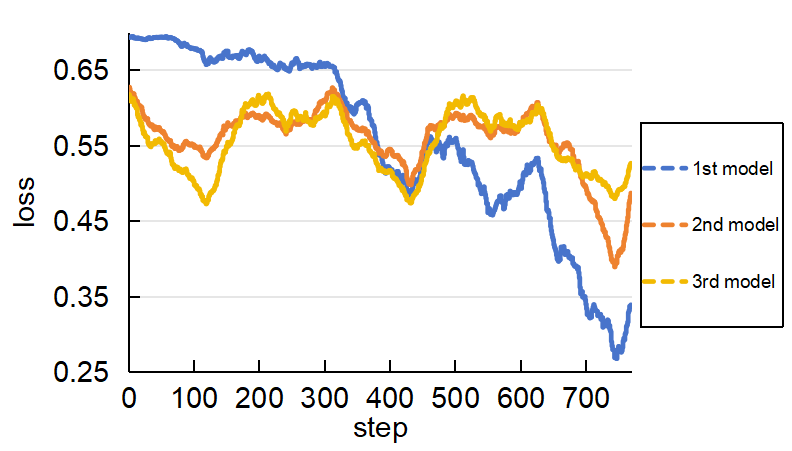}
    \caption{The training loss of three models serially trained on 10,000 samples using Adaboost.}
    \label{fig:3-2-c}
\end{figure}

To evaluate the contribution of AdaBoost and Gradient Boost on weak-to-strong generalization, each method underwent three rounds of training on a weighted set before the ensemble. 
Results are shown in Table~\ref{tab: boost train set}.
It is shown that the boosting methods marginally enhance weak performance, but fail to contribute to W2SG performance, which in some cases, even deteriorated. Investigating the cause, we focus on the training process of the first three models in AdaBoost, as exemplified in Figure~\ref{fig:3-2-c}.
Beyond the initial model, subsequent models struggle to reduce loss, concentrating on challenging samples due to reweighted training, and failing to capture accurate dataset characteristics. 
Thus, the weak performance gain is attributed to a minority of difficult samples, which yet deviate from the overall distribution and do not benefit the complex tasks like knowledge-based Q\&A in W2SG. 
This reaffirms the observations from Section 3.1: in addition to enhancing weak supervision, future efforts should explore more factors to effectively contribute to W2SG.

\section{Improving Weak-to-Strong Generalization with Scalable Oversight}
\label{sec-4}
Scalable oversight leverages auxiliary models to enhance the supervision quality. In this section, we utilize un-tuned strong models as auxiliary models and explore their potential in scalable oversight settings. Two types of auxiliary settings are examined: human-AI interaction and AI-AI debate.

The task remains focused on the SciQ dataset. When selecting models, considering the need for auxiliary models to interact with humans and provide context for the weak models, it's essential that these models possess a certain level of instruction-following and context-understanding capabilities. Taking into account performance on the SciQ task, we chose the Qwen series: qwen-1.8b-chat as the weak teacher model and qwen-7b-base as both the strong student model and the auxiliary model. 

There are two points to clarify: (1) Unlike previous section where ensemble learning leveraged small-sized \emph{base} models trained with task-specific data to serve as weak teachers, in this section we directly employ the small-sized \emph{chat} models as the weak teacher. Since the aim is to verify whether scalable oversight contributes to enhancing annotation quality and subsequently improves weak-to-strong generalization, we do not further train the chat model with task-specific data and directly compare the annotation quality before and after applying scalable oversight.
(2) In typical scalable oversight, it is necessary to tune an auxiliary model according to the assistant task. However, for simplicity in this exploratory study, we don't tune the strong base model to better interact between humans or models. Instead, we utilize in-context learning to ensure a certain level of instruction following, which is also one of the notable limitations of our current approach, awaiting future refinement to fully realize the potential of scalable oversight.
%%宇航
\subsection{Interaction-enhanced Weak-to-Strong Generalization}
\label{sec-4.1}
\begin{algorithm}[t]
\caption{Interaction-enhanced Scalable Oversight.}
\begin{algorithmic}[1] 
\State \textbf{Input:} Unlabeled question-answer dataset $D = \{(q_i, a_i)\}$, size of context example set $m$, number of context examples used per iteration by weak model $n$.
\State \textbf{Output:} Labeled dataset $D_{so}$  through interaction-enhanced scalable oversight.
\State Initialize auxiliary model $M_{aux}$ and weak model $M_{weak}$.
\State Initialize dataset $D_{so} \gets \emptyset$.
\State Initialize context example set $E \gets \emptyset$.

% \Comment{Step 1: In-context example generation for interacting}
\Statex $\triangleright$ Step 1: In-context example generation for interacting.
\For{$i \gets 1$ to $m$}
    \State Randomly select a question $q_i$ from $D$.
    \State Use $M_{weak}$ to generate contextual information $k^w_i$ for $q_i$.
    \State Add $(q_i, k^w_i)$ to the context example set $E$.
\EndFor

% \Comment{Step 2: Scalable oversight for labeling}
\Statex $\triangleright$ Step 2: Scalable oversight for labeling.
\For{each pair $(q_i, a_i)$ in $D$}
    \State Generate augmented background knowledge $k^a_i$ using $M_{aux}$ and $n$ context examples randomly selected from $E$.
    \State Determine the correctness of $a_i$ for $q_i$ as $\hat{y}_i$, based on $k^a_i$.
    \State Add the triplet $(q_i, a_i, \hat{y}_i)$ to $D_{so}$.
\EndFor
\end{algorithmic}
\label{alg:interaction-oversight}
\end{algorithm}

\vspace{2mm}\textbf{Method}\par \hspace{3mm}
%基于问题特点，让辅助模型提供背景知识，然后让弱老师模型以背景知识为context进行回答
%specifically, 由于辅助模型使用的是base，为确保输出背景知识的质量，采用icl控制答案格式
The effectiveness of using an auxiliary model in scalable oversight largely hinges on the specific characteristics of the task.
Considering the SciQ dataset, which includes questions covering a broad spectrum of knowledge across various disciplines such as physics, biology, and earth science, a notable challenge arises for the weak teacher model due to its restricted ability to offer effective annotations.
Therefore, we treat the auxiliary model as a supportive agent to provide essential background knowledge relevant to the questions in the SciQ dataset.
Leveraging this augmented knowledge, the weak teacher model is expected to extend its supervision capability and produce annotations more accurately. \par
Our method's algorithm is detailed in Alg.~\ref{alg:interaction-oversight} which comprises two primary steps.
Since the auxiliary model is the \emph{base} version,
we generate examples in the first step and exploit in-context learning to elicit its instruction following capability.
% Specifically, 
% we randomly select $m$ questions from dataset $D$ and employ the weak teacher model to generate background knowledge responses, as illustrated at the top of Tab~\ref{tab: prompt template}.
% The context example set $E$ is constructed using these question and knowledge pairs.
Note that the basic idea here is to use in-context examples to convey the assistant task's objective and standardize the output format. It is thus less important to guarantee the quality of the generated knowledge responses. Hence, the weak teacher model is sufficient for generating these in-context examples.
In the second step, the auxiliary model initially generates background knowledge related to the posed question. This process involves using $n$ random in-context examples from set $E$. Following this, the weak model judges the correctness of the answers in relation to the question, utilizing the background information provided. 
The corresponding prompt template for the above in-context example generation, knowledge extraction and data annotation are presented in Table~\ref{tab: prompt template} in Appendix. 
% -------- 中文注释 start ---------
% Fellow openai w2s的设置， 把原数据集 sciq 转换为二分类问题。
% We convert each dataset to a binary classification problem. For multiple-choice datasets, suppose each datapoint has a question Q and multiple candidate answers A1, . . . ,Ak. We then convert this datapoint to k new datapoints of the form (Q,Ai), where the label is 0 for all incorrect answers Ai and 1 for the correct answers. 
% 对于sciq 数据集来说，其问题涉及多种学科知识，例如：物理学，生物学，地球科学等， 由于弱老师模型的知识水平有限，所以单独的弱老师模型不能很好的给出标注。 
% 针对这个问题，我们引入了一个辅助模型，通过弱老师模型与辅助模型的交互以获取回答问题所需背景知识，弱老师再利用背景知识给出标注。
% 具体来说，一共分为三步：
% 第一步：弱老师模型询问辅助模型背景知识。 由于辅助模型是base模型，不具备一般的指令跟随能力。 一种解决思路是通过利用 LLM 的icl能力来交代任务意图并指定输出格式。所以我们另弱老师模型针对训练集的随机 m=5 个问题生成回答，构成上下文示例。 接着，对于每个问题我们从构造的m个示例中每次随机选择两个放到辅助模型的 context 中，提示辅助模型生成问题相关的背景知识
% 第二步：弱教师模型利用背景知识标注数据。 将第一步获得的背景知识提供给弱教师模型，弱教师模型综合背景知识给出标注.基于交互式可扩展监督数据微调强模型。 
% 模型结构follow openai 设置。In order to adapt our language models to the classification setting, we replace the unembedding layer of the model with a linear classification head with two outputs. 
% -------- 中文注释 end ---------

\begin{table}[t]
\centering
\begin{tabular}{|c|c|c|}
\hline
 % & Weak performance (qwen-1.8b-chat)  & Weak-to-Strong performance (qwen-7b-base) \\
  & \multirow{2}{*}{\makecell{Weak performance \\ (qwen-1.8b-chat)}} & 
    \multirow{2}{*}{\makecell{Weak-to-Strong performance \\ (qwen-7b-base )}}\\
  & & \\
\hline 
before SO  & 0.676 & 0.826 \\ 
\hline
after SO  & \textbf{0.749} & \textbf{0.848}\\ 
\hline
\end{tabular}
\caption{Improving W2SG via interaction-enhanced scalable oversight.}
\label{tab: interaction-enhanced result}
\end{table}

\vspace{2mm}\textbf{Experimental Settings} \par
\hspace{3mm}\textbf{Models.} 
Our experimental setup consists of three models. 
Firstly, we choose qwen-1.8b-chat as the weak teacher model, which has not been trained on datasets related to the specific task. 
For the strong student model, we employ qwen-7b-base, with a key modification: the unembedding layer of the model is replaced by a linear classification head, configured to produce two outputs. 
The auxiliary model in our study is also based on the qwen-7b-base architecture; however, it is utilized without any training or structural adjustments.\par
\hspace{3mm}\textbf{Weak labels.} 
% This section builds upon the dataset described in Section~\ref{sec-3}. 
We employ the interaction-enhanced method to generate responses on the ``train2'' set.
Weak labels are derived from text responses using a rule-based process. 
For example, if the model outputs ``Yes'', it is  assigned a label of $1$. Conversely, a response of ``No'' is tagged with a label of $0$.
To maintain a balanced distribution of weak labels in our training dataset, we randomly sample 2500 instances from each class to form a training set of 5000 samples for training the strong model.
Similarly, for the validation set, we randomly select 600 instances from each class, resulting in a total of 1200 samples.

\hspace{3mm}\textbf{Training hyperparameters.} 
The finetuning of the strong models is conducted over 2 epochs, utilizing a batch size of 32.
We adopt an early stopping strategy, which is based on monitoring the accuracy of the model against weak labels on a separate validation set.\par

\hspace{3mm}\textbf{Evaluation.} 
We report accuracy on the test set which is also balanced to have an
equal number of samples in each class. In particular, random guess performance corresponds to 50\% accuracy.\par
\vspace{2mm}\textbf{Experimental Results}\par
\begin{figure}[t]
    \centering
    \includegraphics[width=0.6 \textwidth]{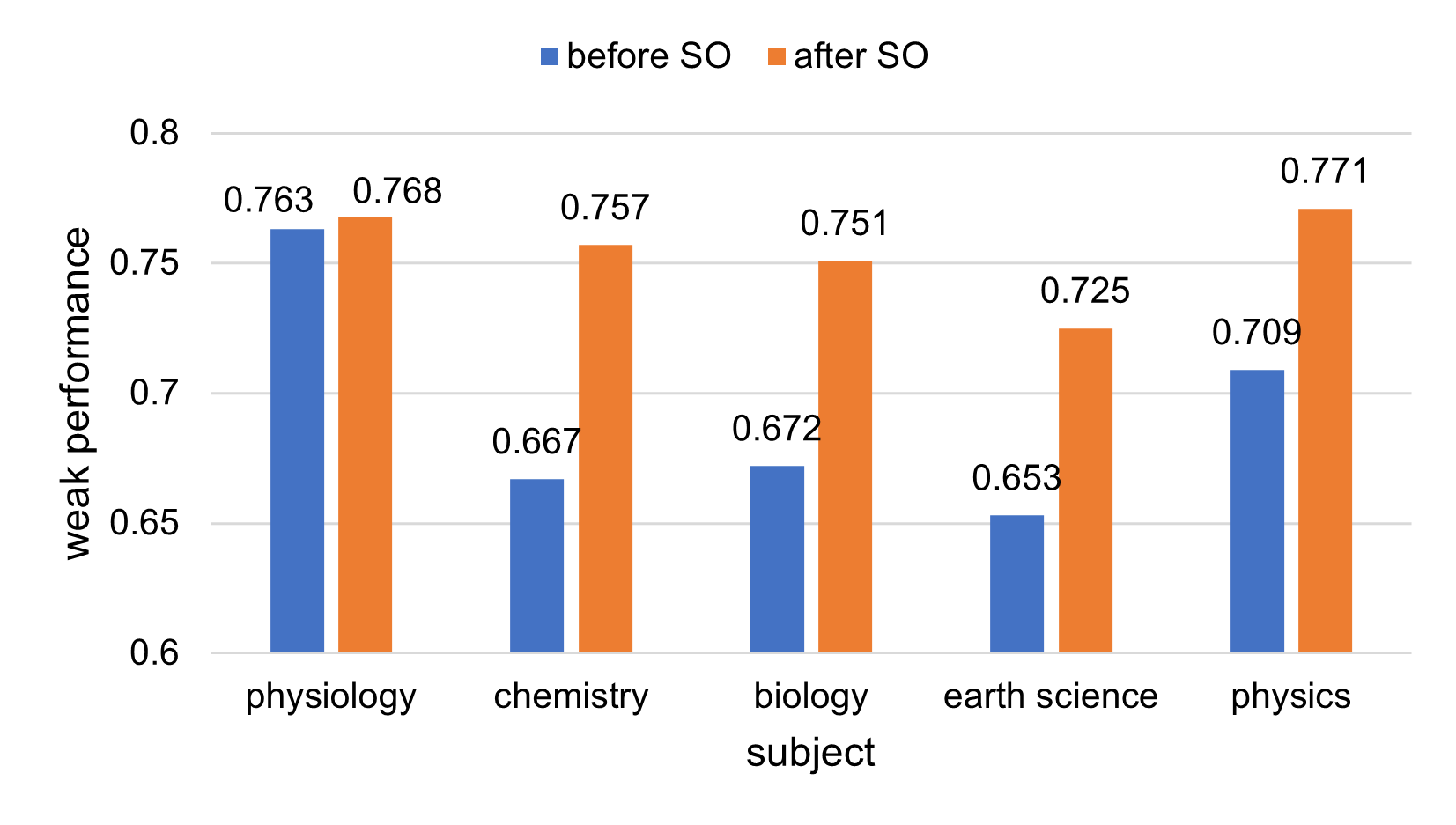}
    \caption{The performance of weak models across different subjects.}
    \label{fig:subject-acc}
\end{figure}

 
We present the experimental results in Table~\ref{tab: interaction-enhanced result}.
Firstly, it's evident that implementing interaction-enhanced scalable oversight substantially enhances the efficacy of weak models, leading to an accuracy increase exceeding 7\%.
Moreover, as demonstrated in Figure~\ref{fig:subject-acc}, we  compare the accuracy across various topics featured in the questions of the test set.\footnote{Note that since the original dataset do not provide topic information, we categorize the questions using ChatGPT-4.}
It shows significant performance improvements in four out of five subjects, with chemistry seeing particularly notable gains.
Such results underscore the crucial role of the auxiliary model's knowledge in tackling the SciQ task, echoing the importance of selecting appropriate assistant tasks. \par
Additionally, the result of the weak-to-strong performance, presented in the last column of Table~\ref{tab: interaction-enhanced result}, 
clearly shows that the proposed method effectively improves weak-to-strong performance. 
This finding further demonstrates that enhancing the weak model's supervisory capabilities through scalable oversight can contribute to the performance of the strong student model.

% \begin{table}[ht]
% \centering
% \begin{tabular}{|c|c|c|}
% \hline
%  & \multirow{2}{*}{\makecell{Weak performance \\ (Qwen-1.8b-chat)}} & \multirow{2}{*}{\makecell{Weak-to-Strong performance \\ (Qwen-7b-base )}}  \\ 
%  & & \\ \hline
% before SO  & 0.676 & 0.765 \\ \hline
% after SO  & 0.749 & 0.796 \\\hline
% \end{tabular}
% \caption{Improving W2SG and interaction-enhanced scalable oversight.}
% \label{tab-1}
% \end{table}

% \begin{table}[ht]
% \centering
% \begin{tabular}{|c|c|c|}
% \hline
%  & \multirow{2}{*}{\makecell{Weak performance \\ (Qwen-1.8b-chat)}} & \multirow{2}{*}{\makecell{Weak-to-Strong performance \\ (Qwen-7b-base )}}  \\ 
%  & & \\ \hline
% before SO  & 0.729 & - \\ \hline
% after SO  &  0.769 & - \\\hline
% \end{tabular}
% \caption{soft label}
% \label{tab-2}
% \end{table}

% \begin{table}[ht]
% \centering
% \begin{tabular}{|c|c|c|c|}
% \hline
%  & \multirow{2}{*}{Weak performance } & \multicolumn{2}{|c|}{Weak-to-Strong performance} \\ 
% \cline{3-4}
%  & & Xent loss & Logconf loss\\ 
% \hline 
% before SO  & 0.676 & 0.765 & - \\ 
% \hline
% after SO  & 0.749 & 0.796 & -\\ 
% \hline
% \end{tabular}
% \caption{Improving W2SG and interaction-enhanced scalable oversight.}
% \label{tab-1}
% \end{table}

% \begin{table}[ht]
% \centering
% \begin{tabular}{|c|c|c|c|}
% \hline
%  & \multirow{2}{*}{Weak performance } & \multicolumn{2}{|c|}{Weak-to-Strong performance} \\ 
% \cline{3-4}
%  & & Xent loss & Logconf loss\\ 
% \hline 
% before SO  & 0.729 & 0.771 & - \\ 
% \hline
% after SO  & 0.769 & 0.793 & -\\ 
% \hline
% \end{tabular}
% \caption{soft.}
% \label{tab-2}
% \end{table}

%%彦谞
\subsection{Debate-enhanced Weak-to-Strong Generalization}
\vspace{2mm}\textbf{Method}\par \hspace{3mm}
%参考SO中经典的debate设置，让强模型之间交互
%我们参考了{Du2023multiagent debate}的设置。具体来说，我们首先实例化两个了Qwen-7B模型作为强老师模型参与辩论。针对SciQ中的一个问答对，这两个强模型分别持有“答案正确”和“答案错误”两个立场，并生成相应的解释。接着，我们将两个强模型的解释通过上下文输入给对方，要求它们根据对方的解释反思并更新自己的解释。如此循环，直到达到预定义的辩论轮数。最后，将最后一轮辩论双方生成的解释同时输入给弱老师模型，即Qwen-1.8B-Chat，使其取代人类裁判的角色，利用两个强模型的解释来判断问题的答案是否正确。（附录中可以配上算法表格和提示模板，也可以画成图片）
The standard debate paradigm involves at least two agents presenting answers and statements to assist human judges in decision-making. In the context of weak-to-strong generalization, auxiliary models engage in the debate, and the weak model replaces humans as the judge. This weak model uses the auxiliary models' arguments for annotating data, which then serves to train the strong student model.

Specifically, our approach consists of three steps, as outlined in Alg.~\ref{alg:debate}. Initially, we deploy two auxiliary models with opposing stances on each question-answer pair from the SciQ dataset. One model affirms the answer's accuracy, while the other challenges it, both providing justifications for their positions. Following the settings in Section 4.1, we utilize qwen-7b-base as our auxiliary models and apply in-context learning to standardize the output formats. 

After generating initial explanations, we conduct several rounds of debate between the two auxiliary agents. The explanation from one agent serves as context for the other, prompting both models to refine their explanations in light of the opposing perspective. We iteratively repeat this multi-agent debate process until a set number of rounds are completed.

Finally, the explanations generated by both agents in the final round of debate are presented concurrently to the weak teacher model. This weak model utilizes these final explanations to assess the answer's validity. The corresponding prompt templates for debate-based scalable oversight are presented in Table~\ref{tab: prompt template debate} in Appendix.
\begin{algorithm}
\caption{Debate-enhanced Scalable Oversight.}
\begin{algorithmic}[1] 
\State \textbf{Input:} Unlabeled question-answer dataset $D = \{(q_i, a_i)\}$, rounds of debate $N$.
\State \textbf{Output:} Labeled dataset $D_{so}$  through debate-enhanced scalable oversight.
\State Initialize two auxiliary models $M_{aux_A}$,$M_{aux_B}$ and weak model $M_{weak}$.
\State Initialize dataset $D_{so} \gets \emptyset$.

%\Comment{Step 1: Initial explanations generation}
\Statex $\triangleright$ Step 1: Initial explanations generation.
\State Initialize dataset $D_1 \gets \emptyset$.
\For{each pair $(q_i, a_i)$ in $D$}
    \State $M_{aux_A}$,$M_{aux_B}$ generate explanations $E^{A1}_i$,$E^{B1}_i$ from opposing sides.
    \State Add $(q_i, a_i, E^{A1}_i ,E^{B1}_i)$ to set $D_1$.
\EndFor
    
%\Comment{Step 2: Multi-agent debate}
\Statex $\triangleright$ Step 2: Multi-agent debate.
\For{$n \gets 2$ to $N$}
    \State Initialize dataset $D_n \gets \emptyset$.
    \For{each pair $(q_i, a_i, E^{An-1}_i, E^{Bn-1}_i)$ in $D_{n-1}$}
        \State Input $E^{Bn-1}_i$ to $M_{aux_A}$ to generate $E^{An}_i$.
        \State Input $E^{An-1}_i$ to $M_{aux_B}$ to generate $E^{Bn}_i$.
        \State Add $(q_i, a_i, E^{An}_i ,E^{Bn}_i)$ to set $D_n$.
    \EndFor
\EndFor

%\Comment{Step 3: Weak model judgement}
\Statex $\triangleright$ Step 3: Weak model judgement.
\For{each pair $(q_i, a_i, E^{AN}_i ,E^{BN}_i)$ in $D_N$}
    \State $M_{weak}$ determine the correctness of $a_i$ for $q_i$ as $\hat{y}_i$, based on $E^{AN}_i$,$E^{BN}_i$.
    \State Add the triplet $(q_i, a_i, \hat{y}_i)$ to $D_{so}$.
\EndFor
\end{algorithmic}
\label{alg:debate}
\end{algorithm}

\vspace{2mm}\textbf{Experimental Settings}\par \hspace{3mm}
The number of debate rounds is set to three. During the first round, both auxiliary models present initial explanations without interaction. In the second and third rounds, the two auxiliary models interact and update their explanations based on each other's statements. After each round, the weak teacher model annotates the dataset based on the arguments presented by both sides. Subsequently, we train the strong student model using the dataset annotated by the weak teacher model. The other settings in this section are identical in Section 4.1. 

\begin{table}[t]
\centering
\begin{tabular}{|c|c|c|}
\hline
 & Weak performance  & Weak-to-Strong performance \\ 
\hline 
before SO  & \textbf{0.676} & \textbf{0.826} \\ 
\hline
after round 1 debate  & 0.642 & 0.819\\ 
after round 2 debate  & 0.662 & 0.822\\ 
after round 3 debate  & 0.667 & 0.822\\ 
\hline
\end{tabular}
\caption{The performance of debate-enhanced scalable oversight.}
\label{tab: debate-enhanced result}
\end{table}

\vspace{2mm}\textbf{Experimental Results}\par \hspace{3mm}
%train2和验证集三轮标注已完成，实验结果待训练完成后补齐。从标注结果来看，随轮数增加标注准确率略有提升，但不明显，且均低于baseline的0.676
%分析：
%-对于SciQ任务，debate设置不适合；
%-弱模型能力不够，无法理解debate设置，同时利用双方信息
%-强学生模型是base，辩论能力不够，需要根据SO设置专门训练assistant AI；
%本节的实验结果如下表所示。首先，可以看到每轮辩论后的weak performance都低于baseline，我们认为有以下3点原因。第一，SciQ是一个基于科学知识的问答任务，答案是客观且确定的，所以在这一任务上设置辩论或许是不适合的。第二，我们所使用的辅助模型是base模型，指令跟随能力较弱，无法充分理解辩论指令，我们用上下文学习来控制输出格式，但较长的上下文增加了辅助模型的理解难度，同时也引入了干扰信息，最终导致生成的部分解释未能符合我们的预期。第三，我们使用的弱老师模型qwen-1.8b-chat，它较小的参数量限制了它的能力，使它在标注数据时难以同时利用正反双方的解释信息。但是，我们也注意到，随着辩论轮次的增加，weak performance提升了，这和{Du2023multiagent debate}的结论是一致的。因此我们相信，若根据debate-enhanced scalable oversight的设置微调辅助模型，同时使用能力更强的弱老师模型，会对Weak-to-Strong Generalization有帮助。
%补：case study
%在表中，我们进行了bad case分析。可以观察到，qwen-1.8b-chat在辩论提示下回答错误的问题，在一般的问答提示下却能够回答正确。这表明过长的上下文以及辩论设置中更复杂的逻辑关系干扰了模型的判断。
The experimental results are shown in Table~\ref{tab: debate-enhanced result}. Despite an observed improvement in weak performance across debate rounds, it does not exceed the results obtained without scalable oversight. This may be attributed to two factors:  Firstly, the auxiliary models we use are base models with limited instruction-following abilities, making it difficult to fully comprehend the debate instructions. Although in-context learning is employed to standardize the output format, the increased length of context proves too challenging, especially for models of 1.8b size. Table~\ref{tab: case study debate} showcases some failure cases. It is observed that qwen-1.8b-chat tends to produce incorrect answers when responding to the \emph{Weak Model Judgement} prompt, yet correctly answers under standard question-answer prompt. This suggests that the extended context and intricate logical dynamics inherent in debate configurations may disrupt the model's ability for accurate evaluation. Secondly, SciQ is a task based on scientific knowledge, where answers are objective and definite. While debate aims at subjective reasoning and may not suit tasks with objective answers, potentially diminishing its utility in this context.

\begin{table}[t]
  \centering
  \begin{tabular}{p{0.45\textwidth}|p{0.45\textwidth}}
    \hline
    \textbf{Debate Prompt} & \textbf{Question-Answer Prompt} \\
     \hline
      Please read the context first and then determine if the following question's answer is correct. If it is correct, reply with solely ``Yes''. If it is incorrect, reply with solely ``No''.\newline
     \textbf{Context}: \emph{One person think it is correct for the reason that} the structure of hydrogen is H, and it is an element that is not always counted as organic, but it is the most abundant element in living organisms. 
    \emph{Another person think it is incorrect for the reason that} organic chemistry is the branch of chemistry that studies the properties and reactions of organic compounds, which are compounds that contain carbon. \newline
    \textbf{Question}: Organic chemistry is the study of the chemistry of what? \newline
    \textbf{Answer}:hydrogen \newline
    \textbf{Output}:Yes.
     & 
      Please determine if the following question's answer is correct.  \newline
      Question: Organic chemistry is the study of the chemistry of what?\newline
      \textbf{Answer}:hydrogen\newline
      \textbf{Output}:The answer ``hydrogen'' is \emph{incorrect}. Organic chemistry is the study of the chemistry of organic compounds, which are molecules composed of carbon atoms bonded to other \emph{non-hydrogen atoms} through chemical bonds such as covalent or ionic bonds. Therefore, the correct answer would be ``organic compounds''.\\
     \hline
      \end{tabular}
  \caption{Example of failure cases in debate-enhanced weak-to-strong generalization.}
  \label{tab: case study debate}
\end{table}

However, it is important to acknowledge the improved weak performance as the number of debate rounds increases, which is consistent with the findings from work ~\cite{du2023improving} on multi-agent debate. More rounds of debate lead to more concise and focused conclusions, thus reducing the evaluative burden on the weak teacher model. Future strategies involving tuning the auxiliary models according to the debate settings and utilizing a more capable weak teacher model are expected to positively contribute to weak-to-strong generalization.

%%宇航
\subsection{Combining Scalable Oversight and Ensemble Learning}
\vspace{2mm}\textbf{Method}\par \hspace{3mm}
Previous experimental observations suggest that scalable oversight and ensemble learning both contribute to enhancing the supervision of weaker models. It is natural to integrate these approaches. Specifically, we combine interaction-based scalable oversight with bagging techniques. Note that since interaction-based scalable oversight requires instruction-following capability, we employ qwen-1.8b-chat as the weak model similar to Section 4.1 and implement two bagging approaches that do not require additional training.
\par
The first bagging approach is akin to the layer-based bagging method introduced in Section~\ref{sec-3}. 
However, in this particular instance, we use the models' original unembedding layer rather than using an additional linear classifier.
The second bagging approach conducts multiple runs of scalable oversight and then amalgamates their outcomes. This process is executed without making any modifications to the model itself. Due to the top-$p$ sampling strategy employed during the generation of responses, different runs exhibit a certain degree of variability which aligns with the conditions for ensemble learning. 

\vspace{2mm}\textbf{Experimental Results}\par \hspace{3mm}
In Table~\ref{tab: weak performance layer result}, we showcase the results of interaction-based SO combing features derived from different layers. 
Note that due to the necessity for interaction, the preceding layers are insufficient for generating coherent natural language responses. The experimental results indicate that the performance of the second-to-last and third-to-last layers is significantly inferior compared to the last layer. Layers preceding these are even incapable of completing the task effectively. This limitation has a consequential impact on the overall performance of bagging, leading to outcomes that are not as optimal as anticipated.\par

\begin{table}[t]
\centering
\begin{tabular}{|l|c|}
\hline
 qwen-1.8b-chat layer  & Weak performance  \\ 
\hline
  last    &  \textbf{0.749} \\ 
 % \hline
  second-to-last   & 0.683  \\ 
% \hline
  third-to-last   &  0.690 \\
\hline
 Bagging (hard voting)   &  0.717 \\
 \hline
\end{tabular}
\caption{The performance of interaction-based SO combined with different layers.}
\label{tab: weak performance layer result}
\end{table}

\begin{figure}[t]
    \centering
    \includegraphics[width=0.69 \textwidth]{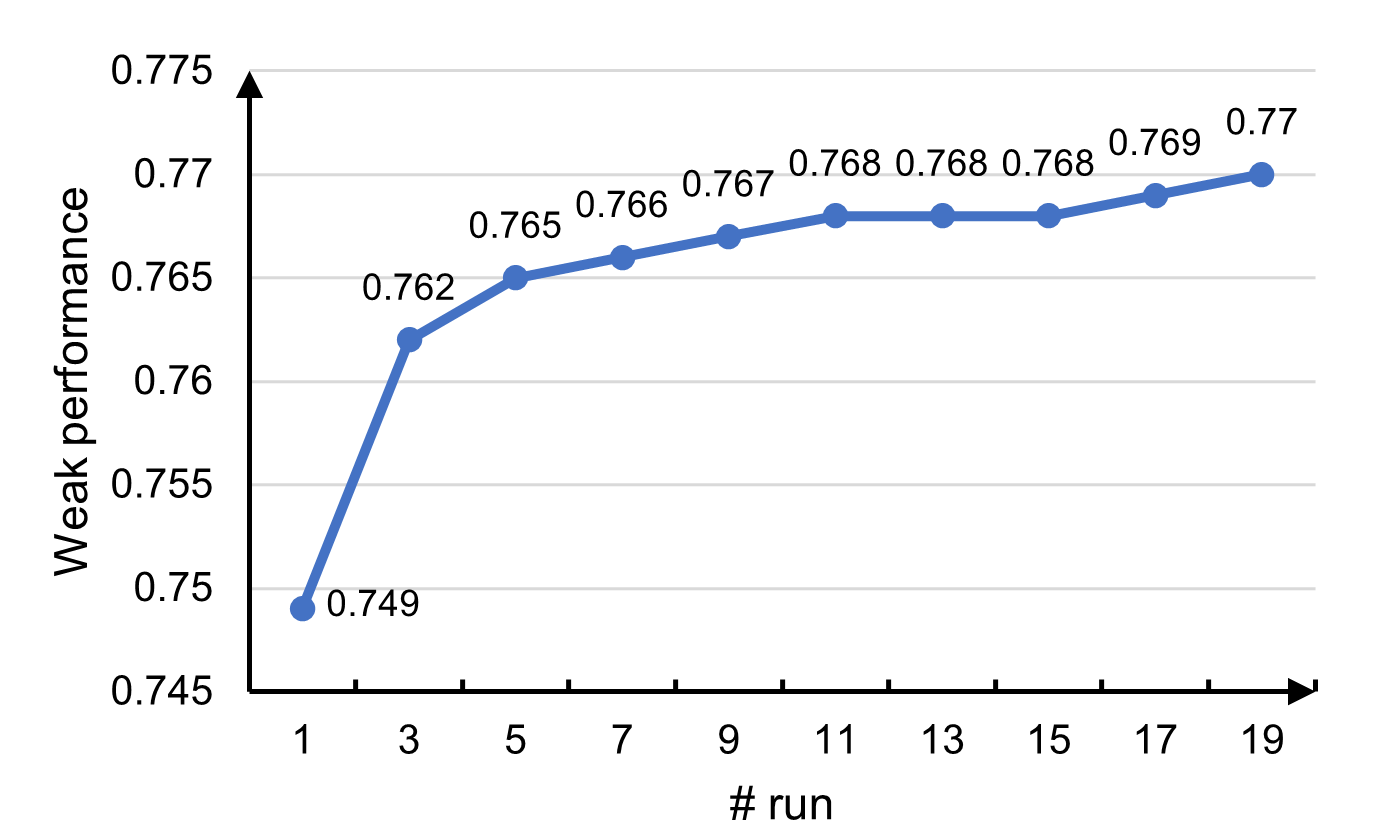}
    \caption{The performance of interaction-based SO combining multiple runs.}
    \label{fig:so+bagging}
\end{figure}

The results of integrating multiple runs of interaction-based SO are depicted in Figure~\ref{fig:so+bagging}. Notably, as we increment the count of runs contributing to the collective inference, the model's effectiveness enhances, reaching a peak performance near 0.77. 
This method is further explored in the weak-to-strong experiment, involving the aggregation of results from 11 different model runs. 
The results, as detailed in Table~\ref{tab: weak2strong SO vote result}, demonstrate that the integration of scalable oversight and ensemble learning contributes to a further improvement in weak-to-strong performance.

It is noteworthy that as the number of runs increases, the improvement in the weak model's performance tends to stabilize, leading to only marginal enhancements in the weak-to-strong approach. 
This phenomenon could be attributed to the limited variability introduced by different runs. 
In future attempts to combine scalable oversight and ensemble learning, it is crucial to not only select an appropriate assistant task but also to strike a balance between ensuring adequate and comparable accuracy and maintaining  diversity  among the weak models. 
One possible approach to achieve this involves utilizing models of comparable size but with varying architectures.
Such a strategy would potentially maximize the benefits of both scalable oversight and the diversity intrinsic to ensemble learning.

 %that when   combining ensemble learning with scalable oversight, it is essential to ensure that each weak model possesses approximately equivalent capacity and is not excessively feeble.
%Secondly, the outputs generated by each model ought to exhibit some variability, for instance, by enhancing the diversity of the samples produced or experimenting with models of comparable size but varying architecture.

\begin{table}[t]
\centering
\begin{tabular}{|c|c|c|}
\hline
 & Weak performance  & Weak-to-Strong performance \\ 
 \hline 
before SO  & 0.676 & 0.826 \\ 
\hline
After SO  &  0.749 & 0.848 \\ 
\hline
After SO + Bagging (\# run = 11)  &  \textbf{0.768} & \textbf{0.851}  \\ 
\hline
\end{tabular}
\caption{W2SG performance combining scalable oversight and ensemble learning.}
\label{tab: weak2strong SO vote result}
\end{table}

%%张京
\section{Discussion: the Contribution of Scalable Oversight to In-Context Learning-based Weak-to-Strong Generalization}
\label{sec-5}
%fine-tune外另一种w2sg设置是ICL；
% -----------start--------------
% "上下文示例"英文表示 contextual examples
% ICL是w2s的另外一种设置。上下文示例的监督信号来自弱模型。本章节我们探索了SO在 improving weak supervision and example selection这两个方面的应用。
In-context learning (ICL) is another configuration of weak-to-strong generalization, where the supervisory signals for contextual examples originate from the weak teacher model. In this section, we explore the applications of scalable oversight in improving weak supervision and selecting similar contextual examples. 
  
% weak superivision; icl-based w2s; 这里SO的思路就是用来提升提升弱监督的质量
% ！！这里记得再润色、逻辑
\subsection{Improving Weak Supervision for ICL-based Weak-to-Strong Generalization}
% \subsection{ ?-enhanced ICL-based Weak-to-Strong Generalization}

\vspace{2mm}\textbf{Method} \par 
% 以往的实验结果证实，相互作用增强的 SO 显著提高了弱模型的监督能力。因此，我们利用SO深入研究ICL范式下的弱到强泛化。
OpenAI's W2SG paper~\cite{burns2023weaktostrong} claims that enhanced weak supervision is also beneficial for the ICL configuration. Given previous findings confirming that interaction-enhanced scalable oversight effectively boosts weak models' supervisory capabilities, we are motivated to explore whether scalable oversight can similarly improve ICL-based weak-to-strong generalization. 

To this end, we design two categories of weak-to-strong prompts: the basic prompt and the confidence-inclusive prompt, as outlined in Table~\ref{tab: prompt icl}. The basic prompt consists of a sequence of question-answer pair examples with pseudo-labels annotated by the weak teacher model. The confidence-inclusive prompt expands on the basic format by incorporating the confidence score associated with these pseudo-labels. Specifically, for the SciQ task, this involves indicating the likelihood (probability of $0$ or $1$) for the first token in the responses from the strong student model.  

%Previous experimental results have confirmed that interaction-enhanced SO significantly enhances the supervisory capabilities of weak models. Consequently, by leveraging scalable oversight, we delve into a comprehensive study of weak-supervision-enhanced ICL-based weak-to-strong generalization.
% 我们设置了两种prompt，标准提示和加入置信度的提示，如表()。标准提示即context是由example的query和对应弱模型给出的伪标签组成。加入置信度的提示在标准提示的基础上，进一步提供了伪标签的置信度。具体到sciq任务中，便是strong model的回答中第一个token对应的“0”或“1”的概率。 
%Specifically, we introduce two categories of weak-to-strong prompts, namely the basic prompt and the confidence-inclusive prompt, as depicted in Table~\ref{tab: prompt icl}. The basic prompt comprises a series of question-answering pair examples and pseudo-labels annotated by the weak teacher model. The confidence-inclusive prompt, an extension of the basic prompt, further integrates the corresponding confidence associated with the pseudo-label. 
% 我们使用5-shot ,并且 为了确保结果统计的稳定性，我们使用strong student model给出的soft label中数值最大应对的标签作为model的答案，统计accuracy。
We implement a 5-shot ICL setup and, to ensure statistical reliability in our results, choose the label with the maximum numerical value among the soft labels generated by the strong student model for accuracy calculation.\par

\vspace{2mm}\textbf{Experimental Settings} \par 
\hspace{3mm}\textbf{Models.} 
% 参照微调的设置，我们使用原始的QWen-1.8B-chat作为weak teacher model，strong student model选择QWen-7B，辅助监督模型亦是如此。
The configuration for scalable oversight aligns precisely with the setup described in Section~\ref{sec-4.1}: we employ qwen-1.8b-chat as the weak teacher model, qwen-7b-base as the strong student model, and use the same setup for the auxiliary model.\par    
\hspace{3mm}\textbf{Sampling Strategy.}  
% 在测试阶段，我们对每个测试样例通过random algorithm获得对应的上下文样例集合。为了避免样例类别的不均衡，我们在选择样例时采用标签均衡的方式。具体来说，当样例总数为k，m个标签类分到的数量将不超过k//m或k//m+1。针对每个test，我们逐一从数据集中采样，每次采样时判断当前soft label所属类别的示例数量是否满足阈值。
Prior research suggests that ICL prediction accuracy is influenced by the proportion of labels among contextual examples~\cite{zhao2021calibrate}. To mitigate this effect, we implement label balancing in the sample selection process. Specifically, for a total sample count denoted by $k$, the quantity of label classes allocated to $m$ will not surpass $k \div m$ for even $k$, or $k \div (m+1)$ for odd $k$. For every test, iterative sampling is conducted from the dataset. During each sampling cycle, we assess if the number of examples within the category matching the current soft label adheres to the predetermined threshold.\par

\vspace{2mm}\textbf{Experimental Results}\par 
% 表()显示了我们的结果。
The resuls of integrating ICL-based weak-to-strong generalization with scalable oversight are shown in Table~\ref{tab:SO in ICL weak performance}. The first finding indicates a notable performance enhancement following this integration. This underscores scalable oversight can enhance generalization performance by improving the accuracy of labels in weak supervision, which aligns with W2SG's findings~\cite{burns2023weaktostrong}. 
% 其次，无论SO与否，conf的加入都对w2s的性能有0.02左右的提升。 confidence可以通过传递更全面的信息在理解准确性方面起着重要作用 。
Secondly, regardless of the presence of scalable oversight, the addition of confidence enhances the performance of weak-to-strong generalization by about 0.02. This shows the pivotal role of confidence in refining accuracy by conveying more comprehensive information. 
% % 相比SO前，conf带来的性能提升是0.0165。而SO后，这个差距有了进一步的增大。所以我们认为SO可以帮助weak model的conf更精准 
Moreover, compared to the base prompt, the inclusion of confidence yields a more substantial impact of weak supervision improvement on weak-to-strong generalization (evidenced by a 0.002 and 0.007 boost for prompts w/o and w/ confidence, respectively). This suggests that scalable oversight not only improves weak supervision accuracy but may also positively influence the refinement of confidence information.\par 
\begin{table}
    \centering
    \begin{tabular}{|c|c|>{\centering\arraybackslash}p{2.8cm}|>{\centering\arraybackslash}p{2.8cm}|} 
    \hline
         & \multirow{2}{*}{Weak performance }  & \multicolumn{2}{c|}{Weak-to-Strong performance} \\ 
         \cline{3-4} &  & w/o confidence & w/ confidence \\  
         \hline   before SO &  0.676& 0.819  & 0.836\\  
         \hline  after SO &  \textbf{0.749} & \textbf{0.821} & \textbf{0.843}\\ 
        % \hline \multirow{2}{*}{10-shot}&  before SO &  0.676 & 0.7293 &  0.801 & 0.832\\ 
        %  &  after SO &  \textbf{0.749} & 0.7611& \textbf{0.8045} & \textbf{0.8345}\\ 
    \hline
    \end{tabular}
    \caption{Improving ICL-based W2SG via interaction-based scalable oversight.}
    \label{tab:SO in ICL weak performance}
\end{table}

\subsection{Selecting Similar Examples for ICL-based Weak-to-Strong Generalization} 
%\vspace{2mm}\textbf{Preliminary Study} \par  
% 影响icl的另一个因素是样例的选择。(cite) 所以我们首先探究了不同的样例选择算法对icl-based w2s的影响。
% Contextual sample selection constitutes another influential factor in ICL~\cite{dong2022survey}. For this purpose, we first explore the impact of different sample selection algorithms on the ICL-based weak-to-strong generalization.  
% % 我们沿用上一节的实验设置，没有SO，使用OpenICL(cite)框架中集成的大部分context retirver算法做了测试。包括bm25, topk, vote。bm25和topk是两个基于相似度的算法，vote的重心在于增加样例的多样性。
% We adopt the experimental settings from the previous section without scalable oversight and conduct tests using several context retriever algorithms integrated into the OpenICL~\cite{wu2023openicl}, including BM25~\cite{robertson2009probabilistic}, Top-K~\cite{liu2021makes}, and VoteK~\cite{su2022selective}. Among them, BM25 and Top-K are two similarity-based algorithms, while VoteK focuses on increasing the diversity of samples.
% % 结果显示在Figure。使用基于相似性的方法可以显著提高性能：基于启发式相似度计算的bm25有超过0.02的提升，之后是另一个重在相似度的topk。
% The results are presented in Table~\ref{tab:icl_retriver}. Performance improvement is most pronounced with the use of similarity-based approaches. TopK, demonstrated an improvement of 0.008, followed by another, BM25 based on heuristic calculation.  
\begin{table}[t]
    \centering
    \begin{tabular}{|c|c|c|c|c|}
    \hline
        ~ & Random & VoteK & BM25 & Top-K \\ \hline
        Weak-to-Strong performance & 0.819 & 0.813 & 0.822 & \textbf{0.827} \\ \hline
    \end{tabular} 
     \caption{The performance of ICL-based W2SG under different context retriever.}
    \label{tab:icl_retriver}
\end{table}

% \begin{figure}[t]
%     \centering
%     \includegraphics[width=0.69 \textwidth]{figs/icl_retriver.png}
%     \caption{The performance of weak-to-strong generalization under different context retriever.}
%     \label{fig:icl_retriver}
% \end{figure} 
%\vspace{2mm}\textbf{Method} \par  
% 在我们的探索中，“相似性”成为最具影响力的方法。因此，我们将其应用于以后的实验。然而，认识到相似性的重要性可能因不同算法而异也是至关重要的。
Contextual example selection is another critical factor in ICL, as highlighted in recent studies~\cite{dong2022survey}. Some works ~\cite{liu2021makes,rubin2021learning} report that selecting similar contextual examples aids ICL effectiveness. To investigate this, we implement various sample selection algorithms for ICL-based weak-to-strong generalization in this subsection.

Specifically, we employ three context retriever algorithms from OpenICL~\cite{wu2023openicl}: BM25~\cite{robertson2009probabilistic}, Top-K~\cite{liu2021makes}, and VoteK~\cite{su2022selective}. BM25 and Top-K are similarity-based, whereas VoteK enhances sample diversity. The results, shown in Table~\ref{tab:icl_retriver}, indicate significant performance improvements with similarity-based methods, particularly Top-K, which achieve a 0.008 enhancement. This underscores the further benefits of selecting similar contextual examples beyond enhancing weak label accuracy for W2SG.

It is crucial to note that while off-the-shelf example selection approaches are used here, determining sample similarity becomes non-trivial for increasingly complex tasks. For instance, gauging similarity in large-scale coding projects is more challenging than evaluation, akin to generation tasks in difficulty. Therefore, a more adaptable and universal approach is to incorporate scalable oversight, utilizing strong auxiliary models for similarity computation and similar contextual example selection. These findings and their implications will be further elaborated in the next paper version.

\section{Conclusions}
The introduction of the weak-to-strong generalization framework has made empirical research in superalignment feasible. However, there remains numerous unresolved conceptual and technical challenges. This study has delineated two phases of superalignment solution under the weak-to-strong generalization framework. In the first phase, we explore two approaches to enhance weak supervision quality. For scalable oversight, the key takeaway is the importance of careful selection of the assistant task and the necessary allocation of the auxiliary model to better fulfill its supporting role. There is also potential in exploring and combining more scalable oversight methods. For ensemble learning, the future focus is on adapting to generative problems and large models, while integrating with other classic machine learning techniques such as weak supervision learning. Moreover, the observed benefits of improved weak supervision for both fine-tuning and in-context learning suggest a combined solution in future weak-to-strong generalization research. In the second phase, preliminary feasibility experiments based on a recursive setup are necessary, along with considering issues of alignment verification and calibration to prevent value drift in the auto alignment evaluator. 

Additionally, it's important to recognize the limitations of weak-to-strong generalization in realistically simulating superalignment scenarios. For instance, in the first phase, the weak teacher models act as proxies for human evaluators. When aligning with superhuman models, the practicality of transitioning these weak models back to human supervisors needs careful consideration. In terms of scalable oversight, methods based on interaction and debate, initially applied to AI models, could be easily adapted for human evaluators. Regarding ensemble learning, there are potential challenges to address: Bagging requires effectively merging diverse viewpoints from evaluators with different backgrounds, while boosting involves coordinating individuals to tackle sequential sub-tasks. Such considerations are crucial for advancing superalignment practices and ensuring their relevance and applicability in future AI development.

%实际操作时，SO辅助任务选择要和要解决的任务适应
%SO效果显著，ensemble learning有限；可能和SciQ任务特点有关

{\small
% \bibliographystyle{ieee_fullname}
% \bibliography{sigproc}
\bibliographystyle{plain} 
\bibliography{main}

\begin{thebibliography}{10}

\bibitem{bowman2022measuring}
Samuel~R. Bowman, Jeeyoon Hyun, Ethan Perez, Edwin Chen, Craig Pettit, Scott
  Heiner, Kamilė Lukošiūtė, Amanda Askell, Andy Jones, Anna Chen, Anna
  Goldie, Azalia Mirhoseini, Cameron McKinnon, Christopher Olah, Daniela
  Amodei, Dario Amodei, Dawn Drain, Dustin Li, Eli Tran-Johnson, Jackson
  Kernion, Jamie Kerr, Jared Mueller, Jeffrey Ladish, Joshua Landau, Kamal
  Ndousse, Liane Lovitt, Nelson Elhage, Nicholas Schiefer, Nicholas Joseph,
  Noemí Mercado, Nova DasSarma, Robin Larson, Sam McCandlish, Sandipan Kundu,
  Scott Johnston, Shauna Kravec, Sheer~El Showk, Stanislav Fort, Timothy
  Telleen-Lawton, Tom Brown, Tom Henighan, Tristan Hume, Yuntao Bai, Zac
  Hatfield-Dodds, Ben Mann, and Jared Kaplan.
\newblock Measuring progress on scalable oversight for large language models,
  2022.

\bibitem{burns2023weaktostrong}
Collin Burns, Pavel Izmailov, Jan~Hendrik Kirchner, Bowen Baker, Leo Gao,
  Leopold Aschenbrenner, Yining Chen, Adrien Ecoffet, Manas Joglekar, Jan
  Leike, Ilya Sutskever, and Jeff Wu.
\newblock Weak-to-strong generalization: Eliciting strong capabilities with
  weak supervision, 2023.

\bibitem{christiano2017deep}
Paul~F Christiano, Jan Leike, Tom Brown, Miljan Martic, Shane Legg, and Dario
  Amodei.
\newblock Deep reinforcement learning from human preferences.
\newblock {\em Advances in neural information processing systems}, 30, 2017.

\bibitem{dong2022survey}
Qingxiu Dong, Lei Li, Damai Dai, Ce~Zheng, Zhiyong Wu, Baobao Chang, Xu~Sun,
  Jingjing Xu, and Zhifang Sui.
\newblock A survey for in-context learning.
\newblock {\em arXiv preprint arXiv:2301.00234}, 2022.

\bibitem{du2023improving}
Yilun Du, Shuang Li, Antonio Torralba, Joshua~B. Tenenbaum, and Igor Mordatch.
\newblock Improving factuality and reasoning in language models through
  multiagent debate, 2023.

\bibitem{liu2021makes}
Jiachang Liu, Dinghan Shen, Yizhe Zhang, Bill Dolan, Lawrence Carin, and Weizhu
  Chen.
\newblock What makes good in-context examples for gpt-$3 $?
\newblock {\em arXiv preprint arXiv:2101.06804}, 2021.

\bibitem{ouyang2022training}
Long Ouyang, Jeffrey Wu, Xu~Jiang, Diogo Almeida, Carroll Wainwright, Pamela
  Mishkin, Chong Zhang, Sandhini Agarwal, Katarina Slama, Alex Ray, et~al.
\newblock Training language models to follow instructions with human feedback.
\newblock {\em Advances in Neural Information Processing Systems},
  35:27730--27744, 2022.

\bibitem{radford2019language}
Alec Radford, Jeffrey Wu, Rewon Child, David Luan, Dario Amodei, and Ilya
  Sutskever.
\newblock Language models are unsupervised multitask learners.
\newblock {\em OpenAI Blog}, 1(8), 2019.

\bibitem{robertson2009probabilistic}
Stephen Robertson, Hugo Zaragoza, et~al.
\newblock The probabilistic relevance framework: Bm25 and beyond.
\newblock {\em Foundations and Trends{\textregistered} in Information
  Retrieval}, 3(4):333--389, 2009.

\bibitem{rubin2021learning}
Ohad Rubin, Jonathan Herzig, and Jonathan Berant.
\newblock Learning to retrieve prompts for in-context learning.
\newblock {\em arXiv preprint arXiv:2112.08633}, 2021.

\bibitem{su2022selective}
Hongjin Su, Jungo Kasai, Chen~Henry Wu, Weijia Shi, Tianlu Wang, Jiayi Xin, Rui
  Zhang, Mari Ostendorf, Luke Zettlemoyer, Noah~A Smith, et~al.
\newblock Selective annotation makes language models better few-shot learners.
\newblock {\em arXiv preprint arXiv:2209.01975}, 2022.

\bibitem{welbl2017crowdsourcing}
Johannes Welbl, Nelson~F. Liu, and Matt Gardner.
\newblock Crowdsourcing multiple choice science questions, 2017.

\bibitem{wu2023openicl}
Zhenyu Wu, YaoXiang Wang, Jiacheng Ye, Jiangtao Feng, Jingjing Xu, Yu~Qiao, and
  Zhiyong Wu.
\newblock Openicl: An open-source framework for in-context learning.
\newblock {\em arXiv preprint arXiv:2303.02913}, 2023.

\bibitem{zhao2021calibrate}
Zihao Zhao, Eric Wallace, Shi Feng, Dan Klein, and Sameer Singh.
\newblock Calibrate before use: Improving few-shot performance of language
  models.
\newblock In {\em International Conference on Machine Learning}, pages
  12697--12706. PMLR, 2021.

\end{thebibliography}
}

\newpage
\section{Appendix}
\subsection{The procedures of Adaboost and Gradient Boost}

% 补充公式等
\begin{algorithm}
\caption{Adaboost}
\begin{algorithmic}[1]
\State \textbf{Input:} training data $D = \{(x_i, y_i)\}$, size of training data $n$  
\State \textbf{Output:} ensemble model $f(x)$
\State Initialize weak model $H_0$
\State Initialize each sample weight $W_0(i) = 1, i = 1,...,n$
\For{$t \gets 0$ to $T$}
    \State Train current weak model $H_t$ with training dataset $D$ and corresponding sample weight $W_t$.
    \State Calculate the error rate $e_t$ of $H_t$ on $D$.
    \State Calculate the $H_t$'s weight $\alpha_t \gets \frac{1}{2}ln(\frac{1-e_t}{e_t})$.
    \State $Z_t \gets 2\sqrt{e_t(1-e_t)}$.
    \For{$i \gets 1$ to $n$}
    \State $W_{t+1}(i) = \frac{W_t(i)exp(-\alpha_ty_iH_t(x_i))}{Z_t}$
\EndFor
\EndFor
\State $f(x) \gets \sum_{i=1}^T\alpha_tH_t(x)$
\end{algorithmic}
\label{alg:adaboost}
\end{algorithm}

\begin{algorithm}
\caption{Gradient Boost}
\begin{algorithmic}[1] 
\State \textbf{Input:} training data $D = \{(x_i, y_i)\}$, size of training data $n$  
\State \textbf{Output:} ensemble model $f(x)$
\State Initialize weak model $H_1$, cumulative model $F_1 = H_1$.
\State Initialize sample gradient $g_1^i = y_i$
\For{$t \gets 1$ to $T$}
    \State $Loss_{t}(g_t, F_t) = \frac{1}{n}\sum_{i=1}^n (g_t^i - F_t(x(i)))^2$
    \State $H_t = argmin_{H_t}Loss_t(g_t, F_t)$
    \For{$i \gets 1$ to $n$}
    \State $g_t^i = \frac{\partial Loss_t(y_i, F_{t}(x_i))}{\partial F_{t}(x_i)}$
    \EndFor
    \State $F_t = F_{t-1} + vH_t(x)$
\EndFor
% \State return $f(x)$
\end{algorithmic}
\label{alg:gradientboost}
\end{algorithm}

\subsection{Prompt Templates}
Table~\ref{tab: prompt template}, Table~\ref{tab: prompt template debate}, and Table~\ref{tab: prompt icl} respectively show the prompts used in interaction-enhanced W2SG (section 4.1), debate-enhanced W2SG (section 4.2) and  ICL-based W2SG (section 5).
\begin{table}[ht]
  \centering
  \begin{tabular}{l p{0.55\textwidth}}
    \toprule 
      & \textbf{Prompt Template} \\ 
    \midrule
     \multirow{3}{*}{\textbf{In-context Examples Generation}} & 
     Please provide the background knowledge to answer the following question. Limit your reply to 30 words.\\
     & Input: [``question''] \\
     & Output:\\
     \midrule
    \multirow{7}{*}{\textbf{ICL-based Knowledge Extraction}}  
     & Question:[``question demo 1'']  \\
     & Knowledge: [``knowledge demo 1''] \\
     & Question:[``question demo 2'']  \\
     & Knowledge: [``knowledge demo 2''] \\
     & ... \\
     & Question:[``question'']\\
     & Knowledge:\\
     \midrule
     \multirow{4}{*}{\textbf{Knowledge-Based Data Annotation}}  
     & Please determine if the following question's answer is correct based on the context. If it is correct, output ``Yes''. If it is incorrect, output ``No''.\\
     & Context: [``knowledge'']\\
     & Input: question:[``question''] answer:[``answer'' ]\\
     & Output:  \\
    \bottomrule
  \end{tabular}
  \caption{The prompt templates used in the interaction-enhanced weak-to-strong generalization.}
  \label{tab: prompt template}
\end{table}

\begin{table}[ht]
  \centering
  \begin{tabular}{l p{0.55\textwidth}}
    \toprule 
      & \textbf{Prompt Template} \\ 
    \midrule
     \multirow{7}{*}{\textbf{Initial Explanation Generation}} & 
     There is a question followed by an answer. Assuming the answer is correct(incorrect), please give your explanation.Examples are given below.\\
    &Input:question:[``question demo 1''] answer:[``answer demo 1'']\\
    &Output:[``explanation demo 1"]\\
    &Input:question:[``question demo 2''] answer:[``answer demo 2'']\\
    &Output:[``explanation demo 2'']\\
    &Input:question:[``question''] answer:[``answer'']\\
    &Output:\\
     \midrule
    \multirow{10}{*}{\textbf{Multi-agent Debate}}  
     & There is a question followed by an answer. Another agent think the answer is incorrect(correct), and its explanation is given below. Please use its explanation as additional information to update your explanation.Examples are given below.  \\
     & Input:question:[``question demo 1''] answer:[``answer demo 1''] \\
     &explanation from another agent:[``another\_explanation demo 1''] \\
     & Output:[``new\_explanation demo 1''] \\
     & Input:question:[``question demo 2''] answer:[``answer demo 2''] \\
     & explanation from another agent:[``another\_explanation demo 2''] \\
     & Output:[``new\_explanation demo 2''] \\
     & Input:question:[``question''] answer:[``answer''] \\
     & explanation from another agent:[``another\_explanation''] \\
     & Output: \\
     \midrule
     \multirow{5}{*}{\textbf{Weak Model Judgement}}  
     & Please read the context first and then determine if the following question's answer is correct. 
    If it is correct, reply with solely ``Yes''. If it is incorrect, reply with solely ``No''.\\
     & Context:One person think it is correct for the reason [``correct\_explanation'']\\
    &  Another person think it is incorrect for the reason [``incorrect\_explanation''] \\
     & Input: question:[``question''] answer:[``answer'' ] \\
     & Output:  \\
    \bottomrule
  \end{tabular}
  \caption{The prompt templates used in the debate-enhanced weak-to-strong generalization.}
  \label{tab: prompt template debate}
\end{table}

\begin{table}[ht]
  \centering
  \begin{tabular}{l p{0.55\textwidth}}
    \toprule 
      & \textbf{Prompt Template} \\ 
    \midrule
     \multirow{9}{*}{\textbf{Basic Prompt}} & 
     There is a science knowledge question, followed by an answer. Respond with 1 if the answer is correct, and with 0 otherwise. \\
    &Q:[``question demo 1''] A:[``answer demo 1'']\\
    &[``label demo 1'']\\
    &Q:[``question demo 2''] A:[``answer demo 2'']\\
    &[``label demo 2'']\\
    &... \\
    &Q:[``question''] A:[``answer'']\\
     \midrule
    \multirow{10}{*}{\textbf{Confidence-Inclusive Prompt}}  
     & There is a science knowledge question, followed by an answer. Respond with 1 if the answer is correct, and with 0 otherwise. Note that there may be errors in the answers to the contextual examples.  \\
     & Q:[``question demo 1''] A:[``answer demo 1''] \\
     &[``label demo 1'']. (Confidence: [``confidence demo 1'']) \\
     &Q:[``question demo 2''] A:[``answer demo 2'']\\
    &[``label demo 2'']. (Confidence: [``confidence demo 2''])\\
    &... \\
    &Q:[``question''] A:[``answer'']\\
    %  \midrule
    % \multirow{8}{*}{\textbf{Weak Model Sample Selection}}  
    %  & Please identify the option that shares the most similar background knowledge with the input. If (A) is similar, output ``(A)''. \\
    %  & A). [``question demo 1''] [``answer demo 1''] \\
    %  & B). [``question demo 2''] [``answer demo 2''] \\
    %  &...\\
    % &Input: [``question''] [``answer'']\\
    % &Output: \\ 
    \bottomrule
  \end{tabular}
  \caption{The prompt templates used in the ICL-based weak-to-strong generalization.}
  \label{tab: prompt icl}
\end{table}

%%%%%%%%%%%%%%%%%%%%%%%%%%%%%%%%%%%%%%%%%%%%%%%%%%%%%%%%%%%%

\end{document}